\documentclass{article}
\usepackage{times}
\usepackage{soul}
\usepackage{url}
\usepackage[hidelinks]{hyperref}
\usepackage[utf8]{inputenc}
\usepackage[small]{caption}
\usepackage{graphicx}
\usepackage{amsmath}
\usepackage{amsfonts} 
\usepackage{amsthm}
\usepackage{amssymb}
\usepackage{booktabs}
\usepackage{algorithm}
\usepackage{algorithmic}
\usepackage{physics}
\usepackage[switch]{lineno}

\usepackage{xcolor}
\usepackage{caption}
\usepackage{subcaption}
\usepackage{bm}
\usepackage{multirow}
\usepackage{booktabs}

\usepackage{authblk}

\title{Do We Need an Encoder-Decoder to Model Dynamical Systems on Networks?}

\date{}

\author[1,2]{Bing Liu}
\author[3]{Wei Luo}
\author[3]{Gang Li}
\author[1,2]{Jing Huang}
\author[1,2]{Bo Yang}
\affil[1]{College of Computer Science and Technology, Jilin University, China}
\affil[2]{Key Laboratory of Symbolic Computation and Knowledge Engineering of Ministry of Education, China}
\affil[3]{School of Information Technology, Deakin University, Geelong, Australia}

\begin{document}

\maketitle

\begin{abstract}
As deep learning gains popularity in modelling dynamical systems, 
we expose an underappreciated misunderstanding relevant to modelling dynamics on networks. 
Strongly influenced by graph neural networks, 
latent vertex embeddings are naturally adopted in many neural dynamical network models.     
However, we show that embeddings tend to induce a model that fits observations well 
but simultaneously has incorrect dynamical behaviours. 
Recognising that previous studies narrowly focus on short-term predictions during the transient phase of a flow, 
we propose three tests for correct long-term behaviour, and illustrate how an embedding-based dynamical model fails these tests,
and analyse the causes, particularly through the lens of topological conjugacy. 
In doing so, we show that the difficulties can be avoided by not using embedding.
We propose a simple embedding-free alternative based on parametrising two additive vector-field components.
Through extensive experiments, we verify that the proposed model can reliably recover a broad class of dynamics on 
different network topologies from time series data.
\end{abstract}

\section{Introduction}

In deep learning, architectural design patterns proven successful for one task are often transplanted 
to tackle related tasks.
One such design pattern is an \emph{encoder-decoder} structure found 
in almost every competitive model, be it a transformer~\cite{vaswani2017attention} or a graph neural network~\cite{kipf2016semi}.
With recent dramatic progress in neural dynamical model \cite{chen2018neural,chamberlain2021grand},
one would naturally assume that an encoder-decoder structure in such models is also helpful and necessary.

We challenge this assumption in the context of 
reconstructing network dynamics from observation time series. 
Many natural and social phenomena
can be modelled as dynamical processes on complex networks, 
such as gene regulatory networks, social networks, and ecological networks~\cite{newman2018networks}.
Recovering the governing mechanism of such dynamical systems from observations 
is the first step towards 
understanding the global dynamical properties and eventually implementing interventions and controls.
In a typical problem setup, a multivariate time series $\bm{x}(t_1), \bm{x}(t_2), \dots, \bm{x}(t_N)$ is observed
from a network dynamical system.
From it, we aim to inductively infer a coupled dynamical system to model $\bm{x}(t)$. 
When the observations are evenly spaced, various solutions exist (e.g., \cite{takens1981detecting,brunton2016discovering}).
But as many complex systems generate unevenly spaced observations, 
recovering the dynamic remains a challenge.

\begin{figure*}[!pt]
    \centering
      \hfill
    \begin{subfigure}[b]{0.45\linewidth}
         \centering
         \includegraphics[width= 0.3\linewidth]{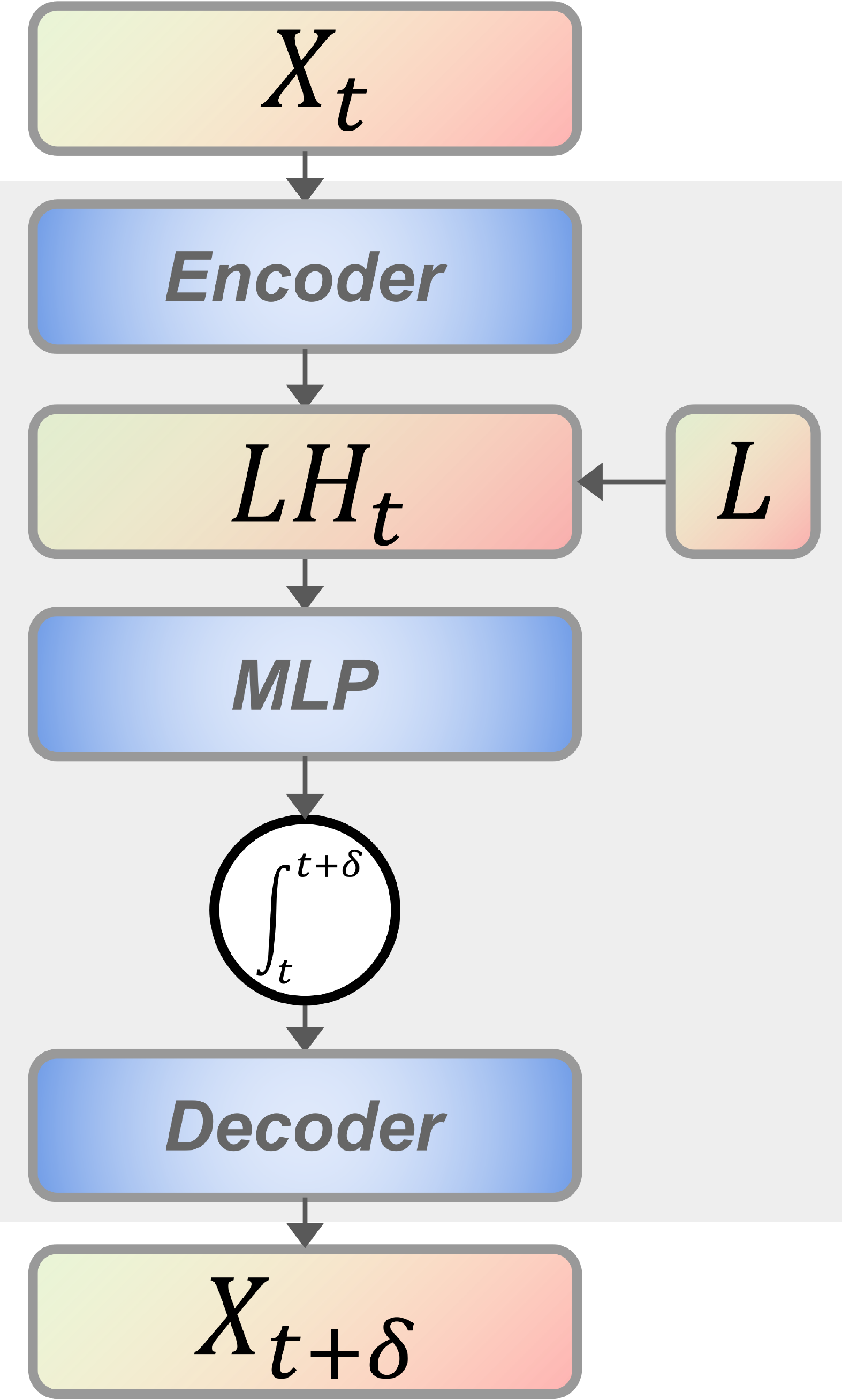}
              \includegraphics[width=0.65\linewidth]{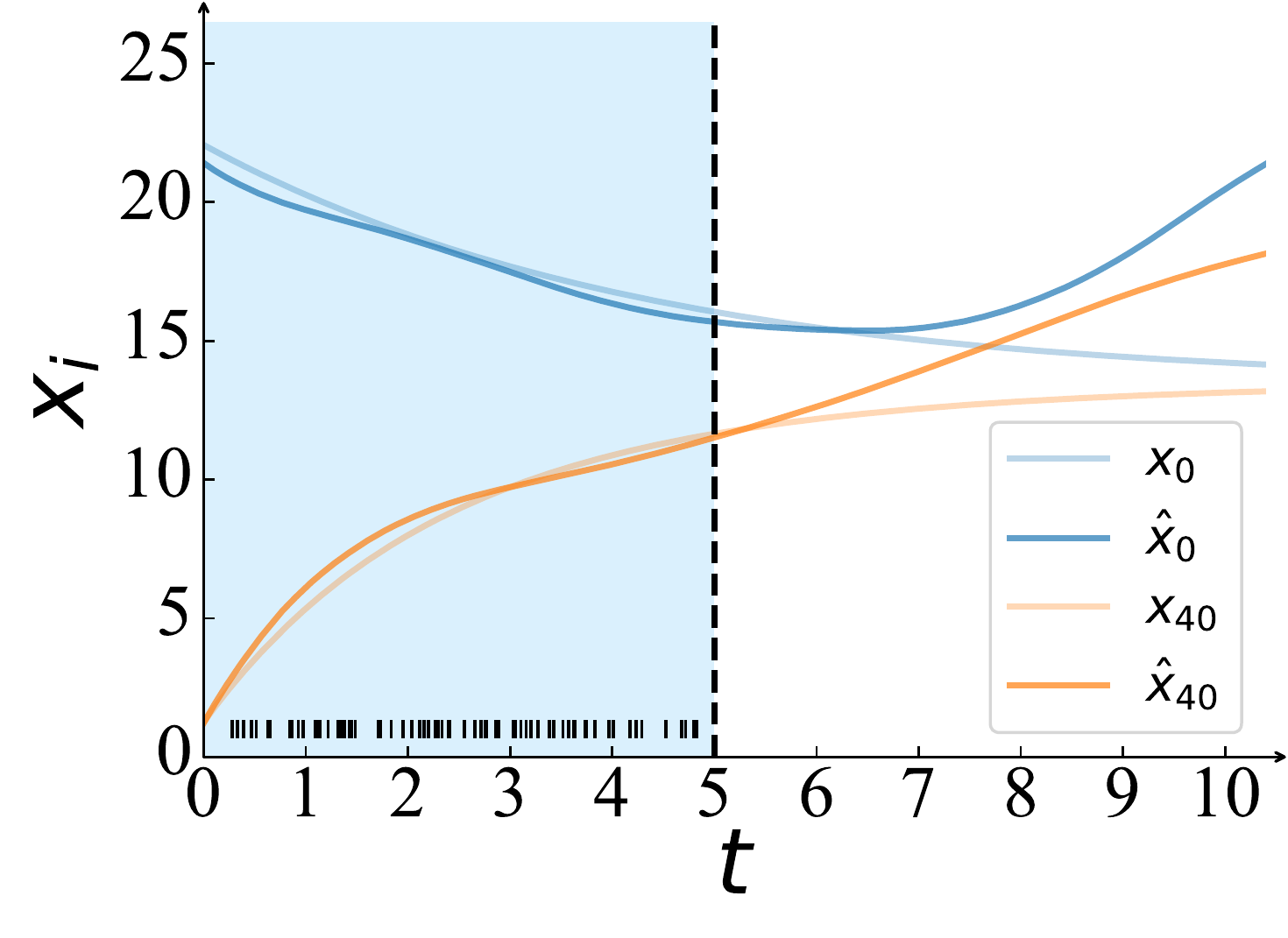}
         \caption{NDCN}
         \label{fig:ndcn}
     \end{subfigure}
     \hfill
     \begin{subfigure}[b]{0.52\linewidth}
         \centering
         \includegraphics[width=0.43\linewidth]{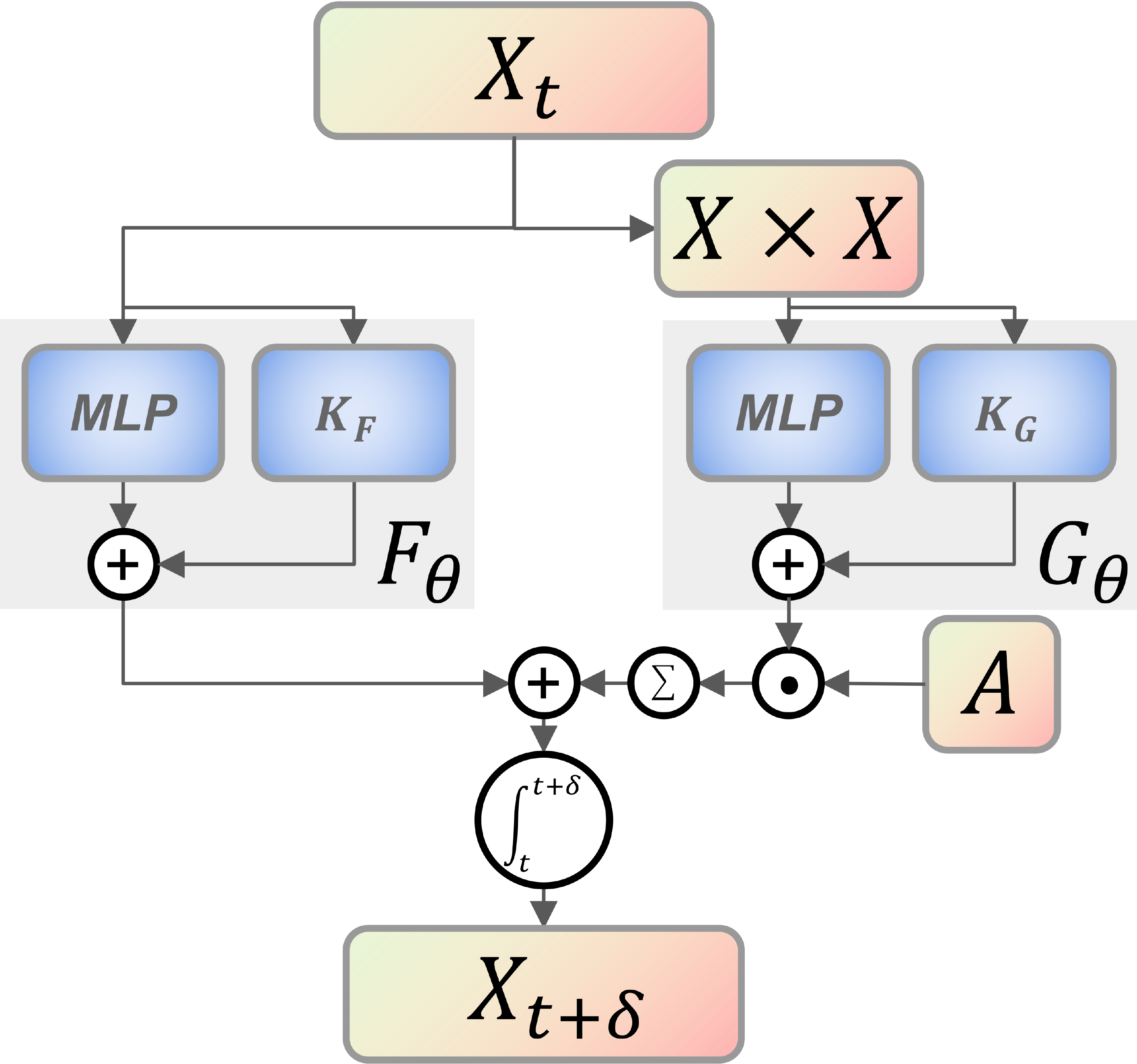}
         \includegraphics[width=0.55\linewidth]{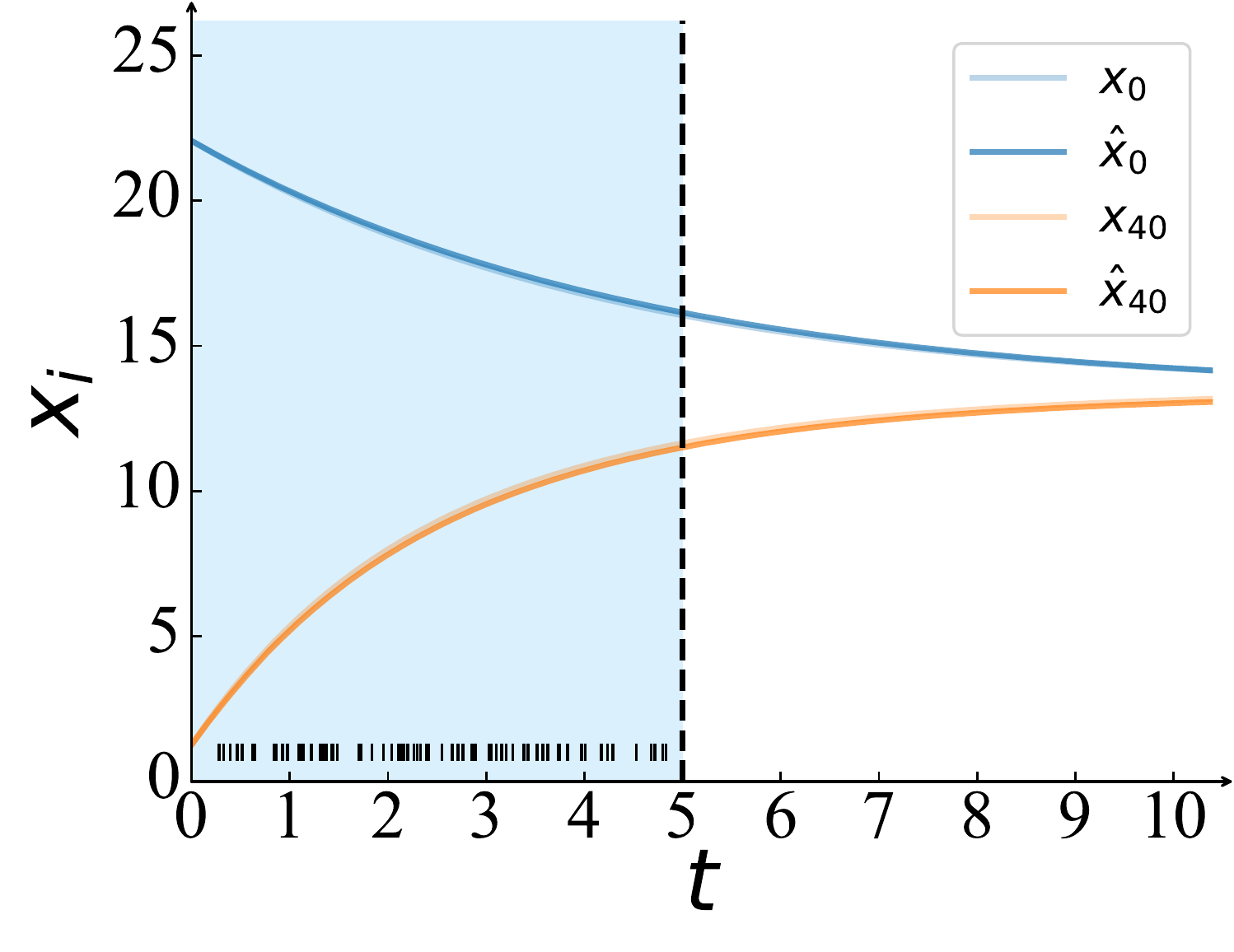}
         \caption{DNND}
         \label{fig:dnnd}
     \end{subfigure}
       \hfill
    \caption{
    Embedding-based neural dynamical model NDCN~\protect\cite{zang2020neural} (a) and our proposed \emph{Dy-Net Neural Dynamics} 
    (DNND) architecture (b).  
    (a) NDCN simultaneously learns an encoder, a decoder, and the evolution rule. 
    The encoder embeds an original variable $x$ into a high-dimensional latent space.
    The evolution rule uses a predetermined linear coupling mechanism (the graph Laplacian)
    to approximate the unknown dynamical interactions among vertices. 
    NDCN suffers from several problems, including poor generalisation beyond the training data 
    (shown as the $x$-$t$ curves diverging from the ground-truth curves from $t>5$).
    (b) In contrast, DNND directly models the network dynamic 
    on the original phase space.
    DNND uses two neural networks $F_\theta$ and $G_\theta$ to separately model the self-dynamic on nodes and the coupling along edges, explicitly allowing nonlinear coupling.
    }
    \label{fig:architecture}
\end{figure*}

Many researchers have already
exploited an encoder-decoder structure to recover coupled flows from time series 
data~\cite{zang2020neural,Yan2021ConTIGCR,Wang2022NDCNBrainAE,Guo2022EvolutionaryPL,Hwang2021ClimateMW}.
Among these, the most representative model is NDCN~\cite{zang2020neural}.
As shown in Fig~\protect\ref{fig:architecture}, 
instead of estimating a flow $\Phi$ from the original observations $\bm{x}$,
NDCN seeks to simultaneously estimate an encoder $f_e$ and a conjugate flow $\Phi'$ such that $\Phi'(t_i, f_e(\bm{x}_0)) \simeq (f_e \circ \Phi)(t_i, \bm{x}_0)$.
As the new flow $\Phi'$ now lives in a higher-dimensional latent space, 
it potentially has a simpler form than $\Phi$, in the same way that kernel methods allow linear algorithms
to be used for nonlinear problems \cite{hofmann2008kernel}.
However, this seemingly innocent use of an encoder can result in an incorrect dynamical model,
in the sense defined by Section~\ref{sec:3_tests}.
In particular, we observe that a flow estimated by NDCN is often dynamically unstable (with the largest Lyapunov 
exponent being positive), even when the time series originates from a stable flow.
We analyse the potential causes.
By concluding that an encoder is unnecessary, we show that it is possible and more desirable to directly
model the dynamics in its original state space, by proposing an effective and versatile embedding-free architecture.

In this paper, we make the following contributions.
\begin{itemize}
\item We expose a severe limitation 
of encoder-decoder-based models for neural dynamics on networks: 
superfluous latent dimensions induce incorrect flows.
In particular, we show that some widely used models do not even produce
a valid dynamical system.
We demonstrate this phenomenon on various dynamics and networks.
\item 
We propose an encoder-free architecture \emph{Dy-Net Neural Dynamics} (DNND)
to learn network dynamics from time series data.
DNND goes beyond mere observation fitting and can reliably learn a model of
the correct dynamical behaviours.
\item We propose a loss warm-up procedure to address the numerical instability in training neural ODE on irregular time series data.
\item 
We improve the existing evaluation protocol for neural dynamic models 
by including metrics aligned with common objectives of analysing nonlinear complex systems, including stability and limiting behaviours. 
\end{itemize}

\section{Background}
\label{AA}

\subsection{Dynamics on Complex Networks}
Consider an undirected network with $n$ vertices.
We represent the network with a symmetric adjacency matrix $\bm{A}$.
An element $A_{ij}$ in $\bm{A}$ is defined as 
\begin{equation}
A_{ij} = \begin{cases}
	1, & \text{if vertices $i$ and $j$ share an edge,}\\
        0, & \text{otherwise.}
\end{cases}
\end{equation}
In this paper, we assume that the network contains no self-loop.
Hence $A_{ii}=0$ for any $i$.

Let $\mathcal{T}\triangleq\mathbb{R}_+$ (or more generally a monoid). 
We assume that each vertex $i$ defines a continuous-time real-value state variable $x_i:\mathcal{T}\rightarrow \mathbb{R}$.
And we consider a dynamical system defined by ordinary differential equations
together with initial conditions:
\begin{align}
\label{eq:mv_odes}
    \dot{x}_i(t)  &=  \Psi_i(x_1(t), x_2(t), \dots, x_n(t))  \\
    x_i(0)  &=  b_i, 
\end{align}
where $t\in \mathcal{T}$ and $1\leq i \leq n$.
In this paper, we use $\dot{x}$ to denote $\dv{t} x$ (i.e., the velocity)
and call $\Psi_i: \mathbb{R}^n \rightarrow \mathbb{R}$ a \emph{velocity operator}.
We assume that each velocity operator is
smooth and globally Lipschitz. 
Also, we restrict $\Psi_i$ to be autonomous (time-invariant).

Let $\bm{x}$ denote $(x_1, x_2, \dots, x_n)^{\top}$.
Then the evolution operators $\Psi_i$ induce a \emph{flow} 
$\Phi: \mathcal{T}\times \mathbb{R}^n \rightarrow \mathbb{R}^n$ as 
\begin{align}
\label{eq:transition_function}
    \Phi(t, \bm{x}(0)) &= \bm{x}(t)  \\
    &= \int_0^{t} \left(\Psi_1(\bm{x}(\tau)),
    \dots,
    \Psi_n(\bm{x}(\tau))
    \right)^{\top} \dd{\tau}
\end{align}

The problem that concerns us in this paper is to recover $\Psi_i$ from a multivariate time series: 
\begin{equation}
    \bm{x}(t_1), \bm{x}(t_2), \dots, \bm{x}(t_N),
\end{equation}    
where $0\leq t_1 < t_2 < \cdots < t_N$. 
Note that we do not assume that the time series is regularly spaced.

Clearly, in its most general form \eqref{eq:mv_odes}, 
it is an ill-defined problem to recover $\Phi$ from a single time series.
If we review $\Delta t_{i} \triangleq t_{i+1} - t_i$ as the value of a roof function $r(\bm{x}(t_i))$,
then $\Phi$ is a special flow~\cite{fisher2019hyperbolic},
and there are infinitely many valid special flows under time changes.
Therefore, we need to introduce an additional inductive bias.
Following~\cite{zang2020neural}, we assume a single operator 
$\Psi:\mathbb{R}\times \mathcal{X}_{\bm{A}} \rightarrow \mathbb{R}$
shared by all state variables:
\begin{equation}
    \label{eq:shared_operator}
      \dot{x}_i(t) =  \Psi(x_i(t), \{ x_j(t) : A_{ij}=1 \}),
\end{equation}
where $\mathcal{X}_{\bm{A}}\subset 2^\mathbb{R}$.
In other words, $x_i(t)$ depends on itself and its neighbouring state variables, 
and $\Psi$ is invariant to the permutation of the neighbouring vertices.

\begin{figure*}[!pt]
    \centering
    \hfill
    \begin{subfigure}[b]{0.23\linewidth}
         \centering
         \includegraphics[width=\linewidth]{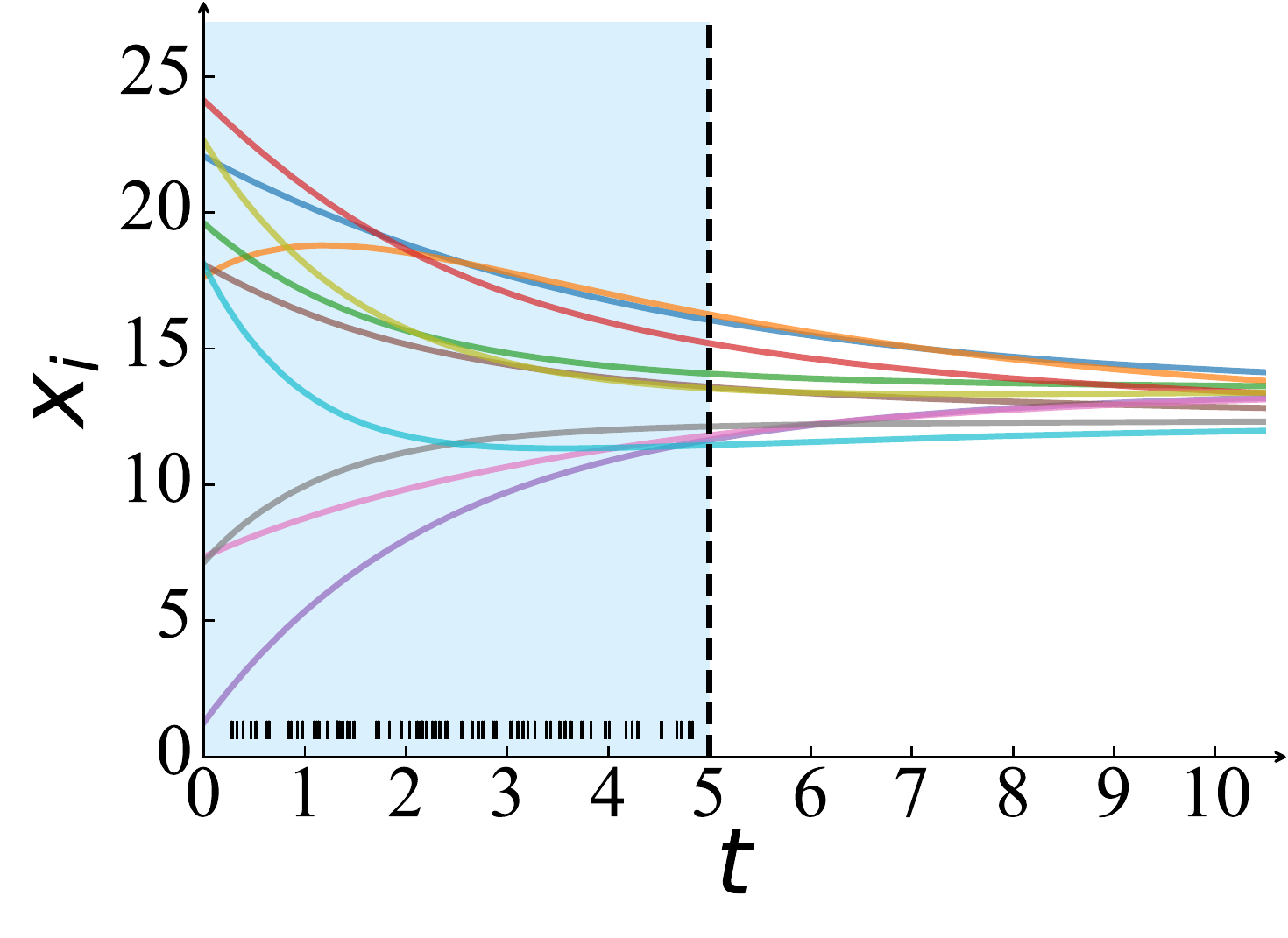}
         \caption{true trajectories}
         \label{fig:true-x-t}
     \end{subfigure}
     \hfill
     \begin{subfigure}[b]{0.23\linewidth}
         \centering
         \includegraphics[width=\linewidth]{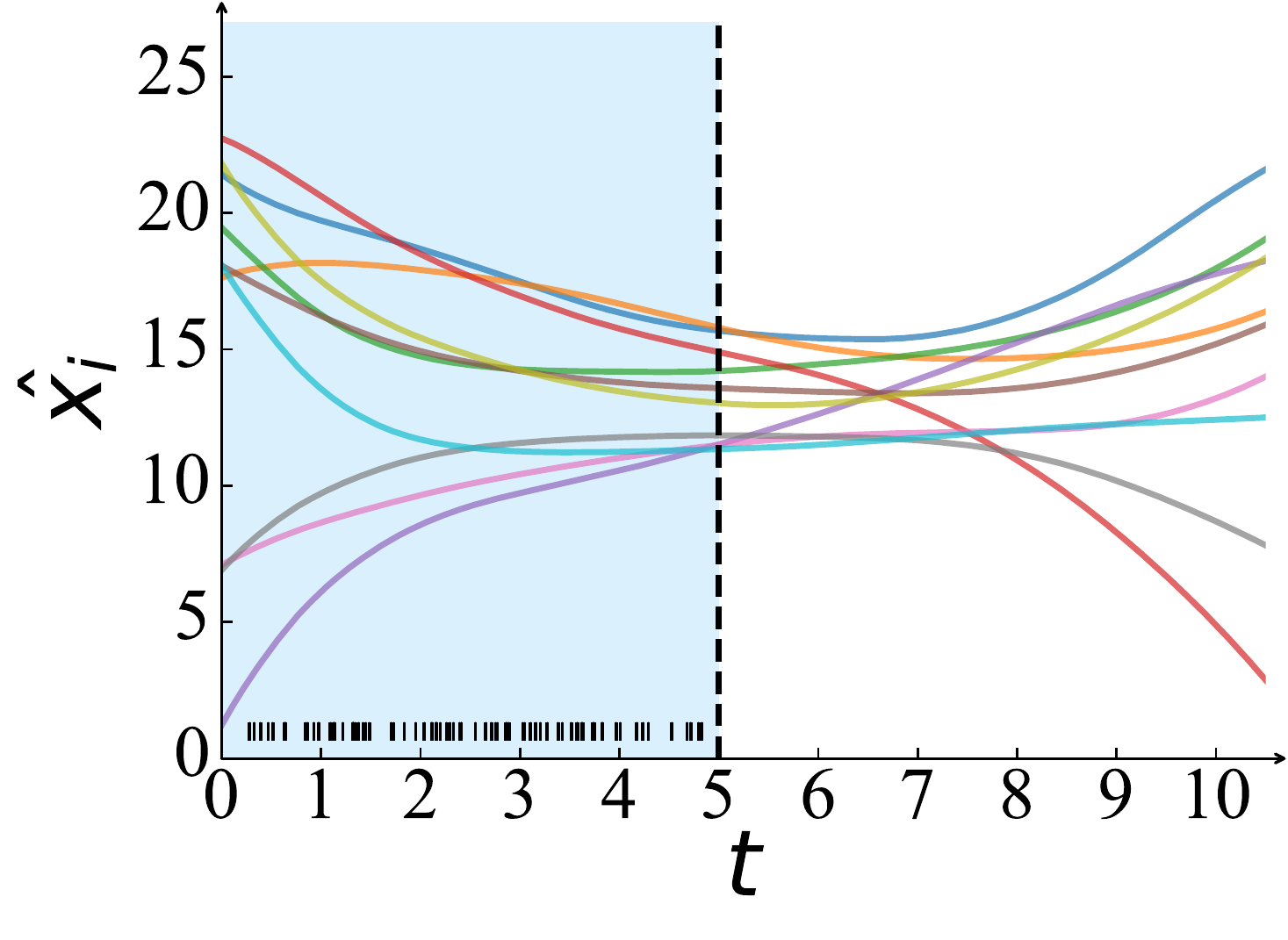}
         \caption{estimated trajectories}
         \label{fig:ndcn-dx-dt}
     \end{subfigure}
     \hfill
     \begin{subfigure}[b]{0.23\linewidth}
         \centering
         \includegraphics[width=\linewidth]{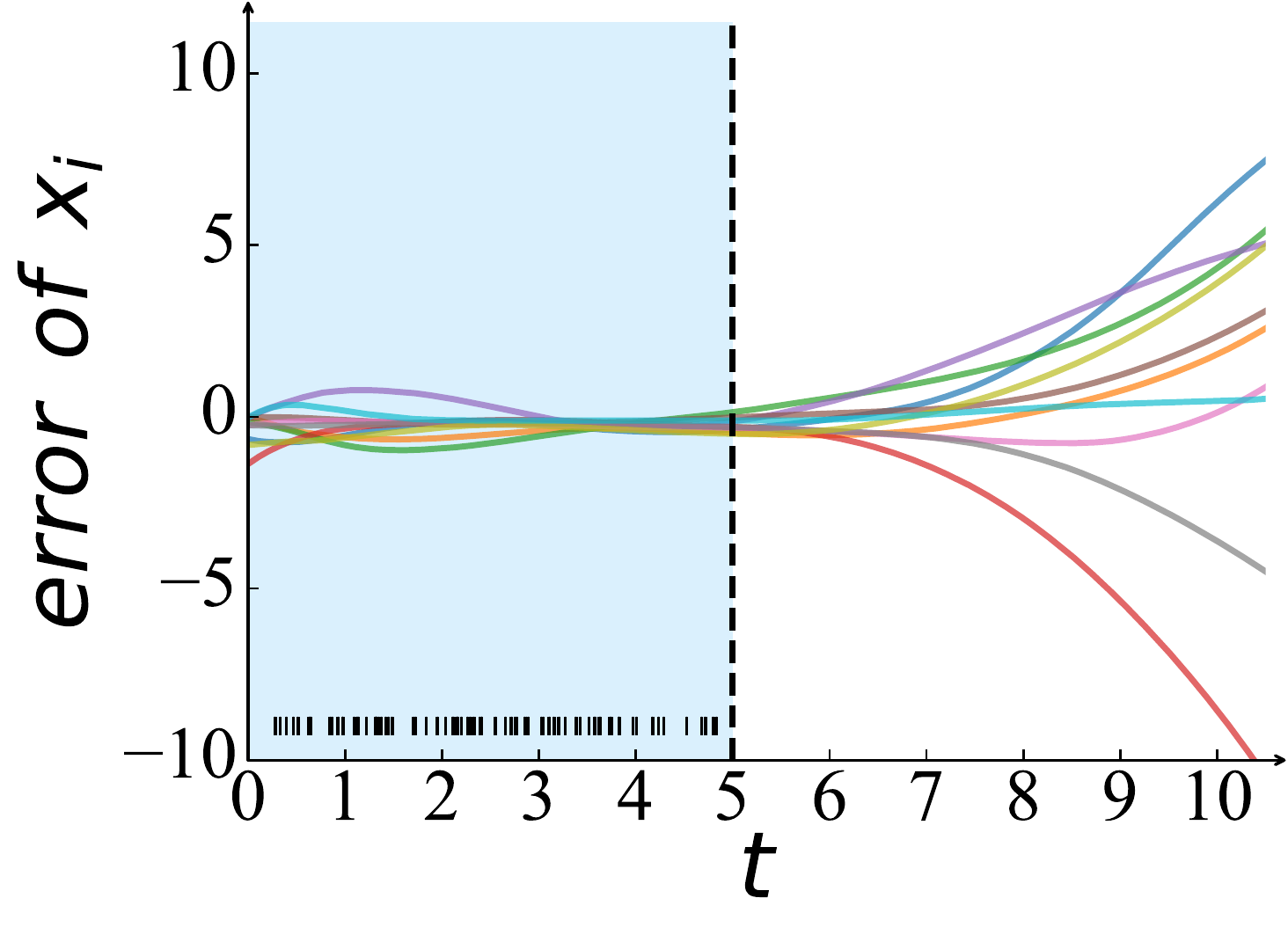}
         \caption{prediction error}
         \label{fig:diff of ndcn and true dx-dt}
     \end{subfigure}
     \hfill 
     \hfill 
     
    \caption{Trajectories of 10 vertices following the heat diffusion dynamics. 
    (a) The ground-truth trajectories converge to the mean temperature. 
    (b) An NDCN model was trained using 80 observations to the left of the dashed line (at times shown in the rug plot). 
    The estimated trajectories fitted the training data well. 
    But the trajectories diverge quickly outside the region with training observations.
    The trajectories certainly are inconsistent with how heat diffusion works.
    }
    \label{fig:ndcn_overfitting}
\end{figure*}

\subsection{Encoder-Decoder Based Neural Dynamical Models}
How do you model the operator $\Psi$ in Eq~\eqref{eq:shared_operator}?
Many neural dynamical models for complex networks (e.g., \cite{zang2020neural}) 
follow a common encoder-decoder architecture in graph neural networks \cite{kipf2016semi} and also in general neural ODEs \cite{lee2021parameterized}.
In NDCN \cite{zang2020neural} (see Fig~\ref{fig:architecture}.(a)), the authors defined
an encoder $f_E: \mathbb{R}\rightarrow \mathbb{R}^d$
and a decoder $f_D: \mathbb{R}^d \rightarrow \mathbb{R}$, 
where $d\in \mathbb{Z}_+$ is the embedding dimension,
so that each initial value $x_i(0)$ is encoded jointly as $d$ state variables in the vector
 $\bm{h}_i(0)=f_E(x(0))$.
Then they define
\begin{equation}
\label{eq:laplacian_embedding}
\dot{  \bm{H}}(t)  = f_{\theta}(\bm{L} \bm{H}(t)),
\end{equation}
where $\bm{L}$ is the graph Laplacian, $\bm{H}(t) = (\bm{h}_1(t), \bm{h}_2(t), \dots, \bm{h}_n(t))^\top$,
and $f_{\theta}:\mathbb{R}^d \rightarrow \mathbb{R}^d$ is a vector-valued function.
Clearly, the use of graph Laplacian for aggregation is a convenient choice. 
It restricts the type of dynamical coupling among neighbouring vertices.

Training an NDCN model involves jointly fitting $f_E$, $f_\theta$, $f_D$ to minimise 
the empirical risk on the observations.
It is not different from a standard supervised training setup. 
In particular, no explicit requirement is imposed on the dynamical behaviours of the fitted flow. 

\section{Issues with Encoder-Decoder Based Neural Dynamical Models}
In this section, we show that encoding the scalar state variable $x(t)$ as a latent vector $\bm{h}(t)$ leads to overfitting and spurious evolution operators.
We use a simulated example to demonstrate the problems 
and then analyse potential causes in Section~\ref{sec:problem_analysis}. 
Later in Section~\ref{sec:dnnd}, we will propose a simple yet effective solution to address these problems.

\subsection{Three Tests for Neural Dynamical Models}
\label{sec:3_tests}

By the universal approximation theorem \cite{hornik1989multilayer},
we can fit a neural network to approximate any time series.
In NDCN, if we keep increasing the embedding dimension $d$,
we can always fit a given observation time series.
But for any fitted neural dynamical model to be useful, 
we expect the induced flow to have the correct dynamical properties,
which we aim to capture with the following three tests.

\paragraph{Extrapolation Beyond Training Data.}
To fit a neural dynamical model, we rely on observations at finite time points 
$t_1, t_2, \dots, t_N$. 
These observations are often collected during a short-term transient period close to an initial condition.

In many studies of NODEs, a test sample from a time period overlapping or close to the training period is used.
This is consistent with the standard supervised setting, where data is assumed to be independent and identically distributed (iid) or nearly iid.

However, in dynamical modelling, more important is the long-term behaviour of the fitted model. 
Therefore, the model needs to make accurate predictions on long-term horizons.
So our first test is \emph{whether the estimated model can make accurate long-term predictions beyond the observation period of the training data.}

\paragraph{Same Fixed Points or Invariant Sets.}
One may argue that in dynamic system modelling, 
what matters is the correct final state, 
instead of accurate predictions of intermediate state values.

Related to the previous test is \emph{whether the fitted model needs to have the same collection of fixed points (or invariant sets) as the true model.} 
If the answer is yes, two corresponding fixed points need to have the same dynamical stability.
In this paper, we check the stability of a flow by the largest Lyapunov exponent~\cite{wolf1985determining}.

\paragraph{Neural Network as a Legal Flow.}
The final test is the most important but is often overlooked: 
\emph{whether the fitted neural dynamical model induces a legal vector field for a flow}.
In particular, the model should induce a state transition function that is a monoid action.
This means that if $\Phi$ is a flow, then 
\begin{align}
\Phi(0, \bm{x}) & = \bm{x} \\
\Phi\left(t_2, \Phi\left(t_1, \bm{x}\right)\right) & =\Phi\left(t_2+t_1, \bm{x}\right),
\end{align}
for any $t_1, t_2\in \mathcal{T}$.
In other words, 
the trajectory $\bm{x}(t)$ cannot branch out at any time $t$.

\subsection{How NDCN Fares in These Tests}
\label{sec:ndcn_problems}

We demonstrate the problems with embedding-based neural dynamical models using an example from \cite{zang2020neural}.
In this example, NDCN is shown to outperform various RNN+GNN baselines, in short-horizon extrapolation tasks  (see Table~3 of \cite{zang2020neural}).

\paragraph{Heat Diffusion on a Grid Network.}
Following \cite{zang2020neural}, 
we simulate a vector time series using a heat equation defined on a grid network.
We consider a network with $400$ vertices arranged in a square 2d lattice.
Let $\{v(k, l) : 1\leq k, l \leq 20\}$ be the vertices.
Two vertices $v(k,l)$ and $v(k',l')$ are connected if and only if $|k - k'|\leq 1$ and $|l- l'| \leq 1$.
In other words, this is a modified version of the $8$-regular graph with 2d boundaries.
On this network, a flow is defined by the heat equation.
For each node $i$, we have: 
\begin{equation}
\label{eq:laplacian_1d}
\dot{x}_{i}(t) =-\sum_{j = 1}^{N} A_{i j} \alpha (x_{i} - x_{j}),
\end{equation}
where $\alpha$ is a diffusive constant.
Also $x_i(0)$ is set with a random value from $[0, 25]$.
Note that in theory, the above coupling operator can be perfectly modelled by the graph Laplacian.

From the above flow, we created $80$ irregularly spaced samples with $t\in [0, 5]$,
and fitted an NDCN model using the code provided by the authors on Git Hub (\url{https://github.com/calvin-zcx/ndcn}).

\paragraph{Failing to Extrapolate.}
Fig~\ref{fig:ndcn_overfitting} shows the true trajectories of $10$ vertices and 
the corresponding trajectories recovered from the NDCN model.
The NDCN model performed well in the neighbourhood of the training observations (to the left of the dashed line).
But it has almost poor generalisation capacity outside that region.
Fig~\ref{fig:ndcn_overfitting}(c) shows that the prediction error grows 
as we move away from the training data.

\paragraph{Missing the Fixed Point.}
The heat flow has a simple fixed point: 
eventually, every $x_i$ converges to the average temperature of the network. 
Fig~\ref{fig:ndcn-dx-dt} shows that 
the NDCN model fails to capture this very stable fixed point, 
as all state variables actually diverge over time in the fitted model.

We measured the largest Lyapunov exponent of the true flow 
and the estimated flow.
While the true value is $-35.83$, indicating a stable flow,
the value of the estimated flow is $9.36$, a positive value that implies instability.

The same contradiction can be revealed by a 2D visualisation of a vertex-wise phase plot.
Fig~\ref{fig:phase_plot} shows that the estimated flow initially approached the average temperature
but eventually diverged from the correct fixed point.

\paragraph{Failing to Produce a Flow.}
From the estimated flow $\hat{\Phi}$, we produce two time series.
One from $t=0$ with the initial condition $x_i(0)$ and another from $t=5$ with the initial condition $\hat{\Phi}(5, x_i(0))$.
If $\hat{\Phi}$ is a legal flow, then these two time series should coincide from $t\geq 5$.

\begin{figure}[t]
    \centering
           \includegraphics[width=0.6\linewidth]{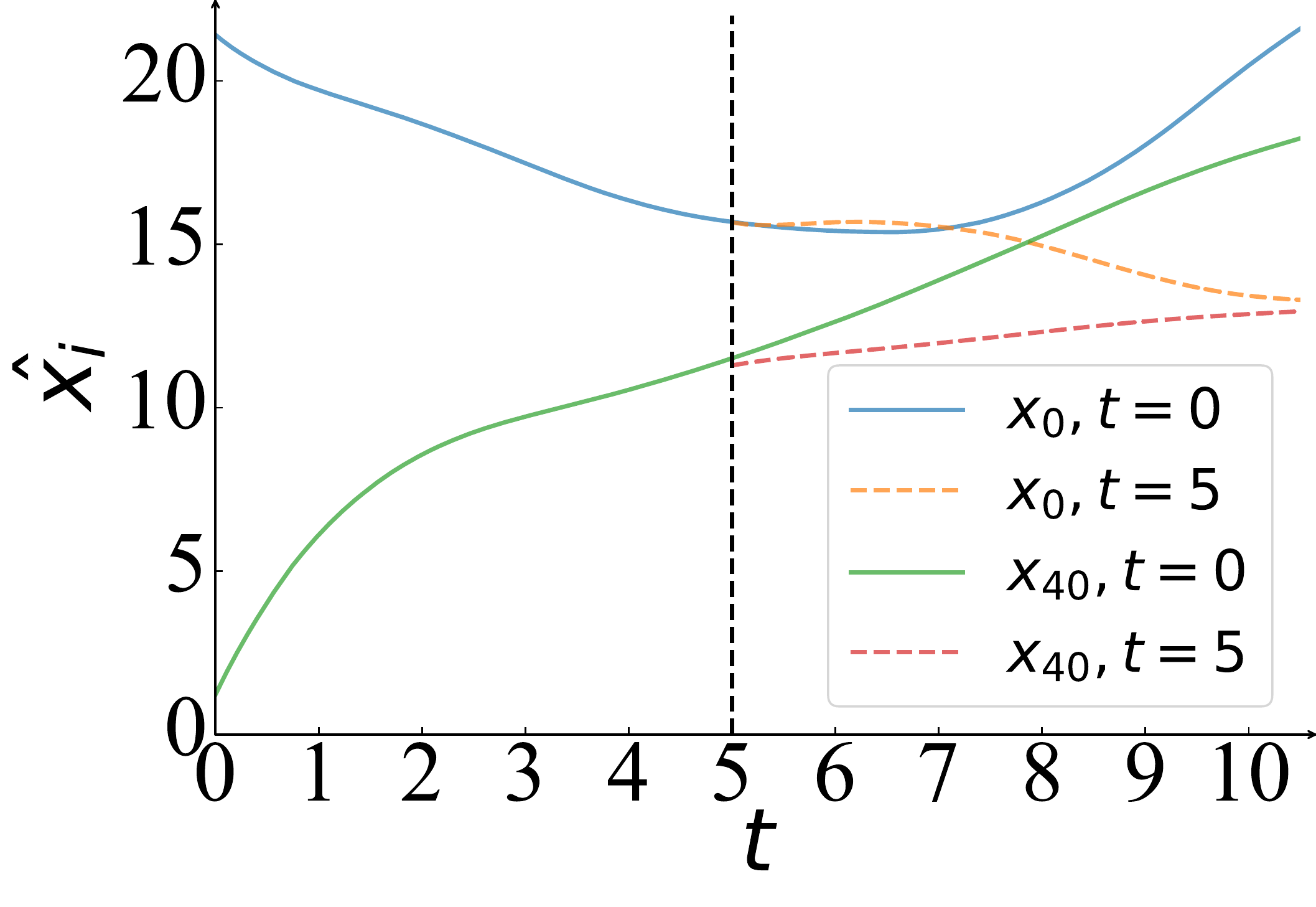}
    \caption{NDCN does not define a proper dynamical system 
         as $\Phi\left(t_2, \Phi\left(t_1, \bm{x}\right)\right) \not =\Phi\left(t_2+t_1, \bm{x}\right)$,
         violating the monoid-action requirement.
         This is shown as one trajectory branching into two trajectories (dashed vs solid). }
    \label{fig:inconsistency in definition}
\end{figure}

Figure~\ref{fig:inconsistency in definition} shows the time series for two vertices.
Clearly resetting the initial condition led to a completely different time series.

\subsection{What Went Wrong?}
\label{sec:problem_analysis}
The above example shows that a neural dynamical model can fit an observation time series well in the embedding space, but at the same time fails to produce the correct flow. 
Embedding has been used to reconstruct flows from regularly spaced time series~\cite{takens1981detecting}.
Also, it has been shown that any homeomorphism $h: \mathcal{X} \rightarrow \mathcal{X}, \mathcal{X} \subset \mathbb{R}^p$ in a p-dimensional Euclidean space can be approximated by a Neural ODE $\hat{\Phi}: \mathcal{T} \times \mathbb{R}^{2 p} \rightarrow \mathbb{R}^{2 p}$  operating on a 2p-dimensional Euclidean space \cite{zhang2020approximation}.
But then why did the embedding-based model fail to recover the heat dynamics?

\paragraph{Encoder/Decoder or Homeomorphism?}
First, we limit our definition of a \emph{correct} flow.
Following \cite{fisher2019hyperbolic}, a flow $\Phi=\{\phi(t, x)\}$ on $X$ and another flow $\Psi=\{\psi(t, y)\}$ on $Y$ are \emph{topological equivalent} if there is a homeomorphism $h: X\rightarrow Y$ such that 
\begin{equation}
    h(\phi(t, x)) = \psi(t, h(x)).
\end{equation}
If we accept the notion of ``correctness", then in Fig~\ref{fig:architecture}(a) the encoder function $f_E$ and the decoder function $f_D$ have to satisfy some conditions.

First, $f_E$ and $f_D$ need to be inverse to each other, as they represent $h$ and $h^{-1}$.
This implies that $f_E$ and $f_D$ should not be implemented using two independent neural networks.
Secondly, both $f_E$ and $f_D$ must be one-to-one functions (injections).
This implies that the encoder/decoder neural network has to be reversible. 
In particular, it cannot contain pooling layers or a RELU activation function.
However, the encoders and decoders used in most neural dynamical models do not meet these two conditions.
In \cite{zang2020neural}, $f_D$ and $f_E$ are parametrised using two separate neural networks, 
just as in a standard GNN.

\begin{figure}[tbp]
\hspace*{\fill}
    \begin{subfigure}[b]{0.3\linewidth}
         \centering
            \includegraphics[width=1\linewidth]{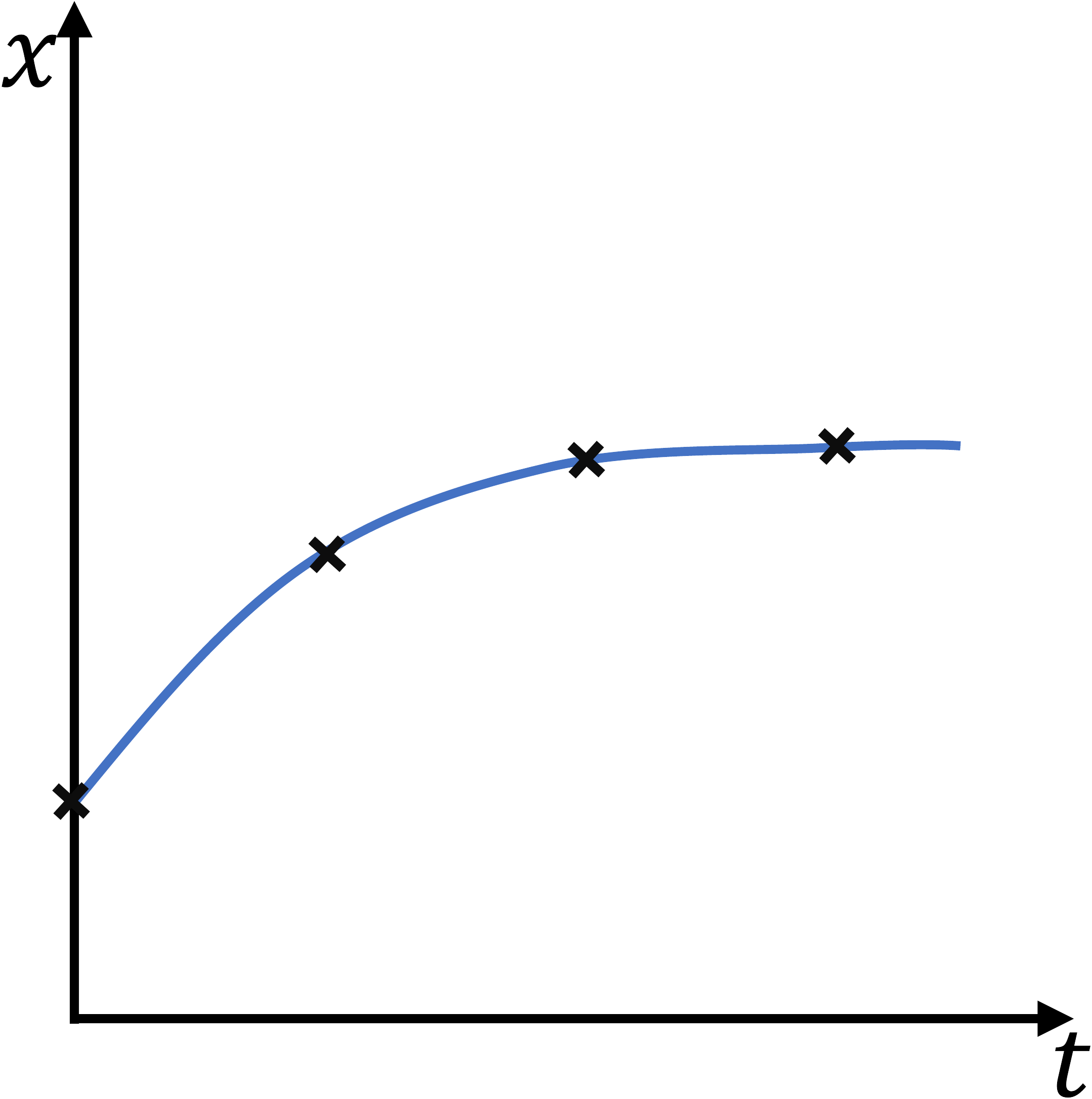}
         \label{origin line}
    \end{subfigure}
\hspace*{\fill}
    \begin{subfigure}[b]{0.35\linewidth}
         \centering
            \includegraphics[width=1\linewidth]{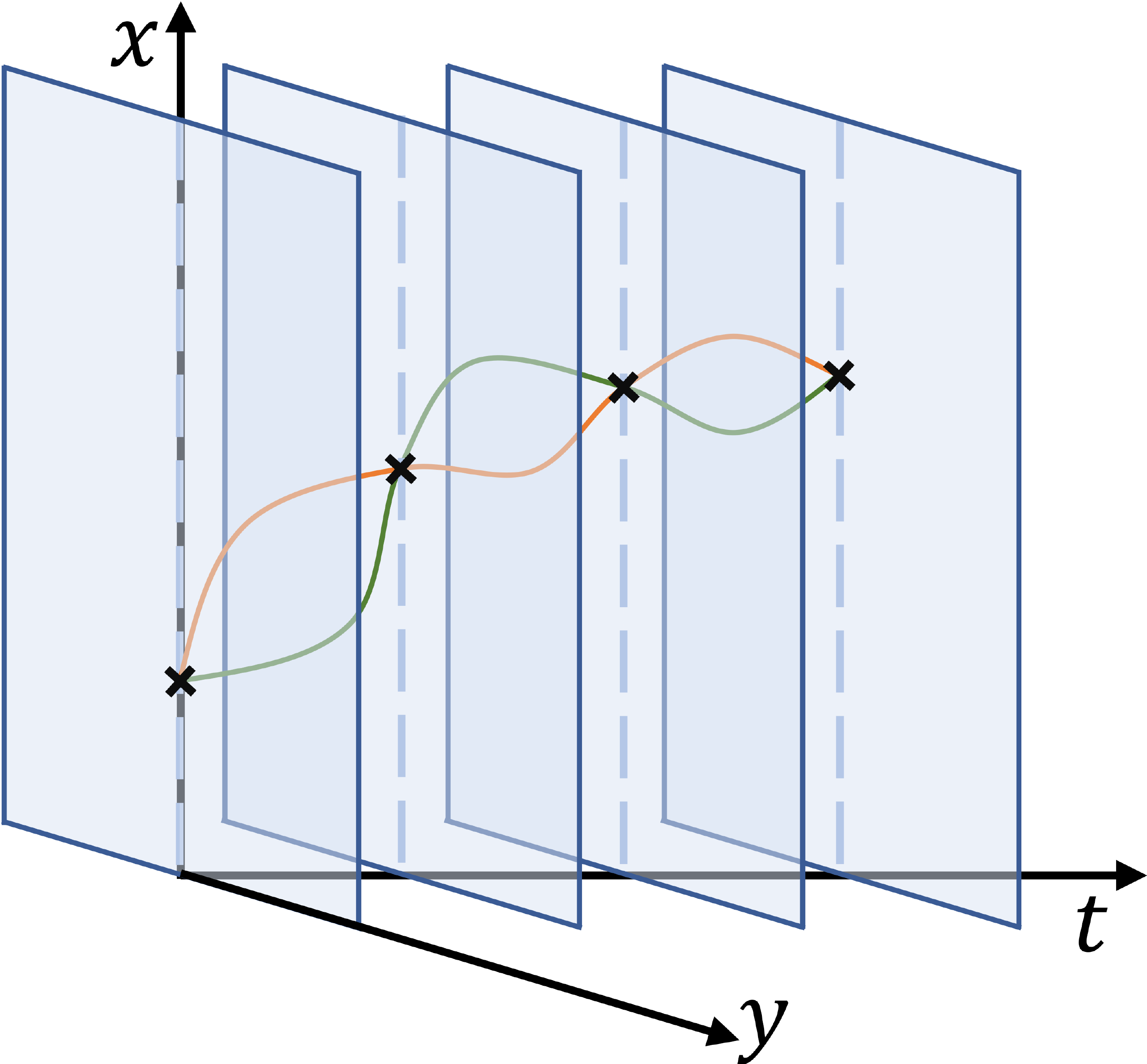}
         \label{encoder line}
    \end{subfigure}
\hspace*{\fill}
\caption{Additional embedding dimensions reduce the influence of observation data on the model fit.
Increased hypothesis space leads to overly complex and unstable models, as suggested by Occam's razor~\protect\cite{blumer1987occam}.}
\label{fig:line_picture}
\end{figure}

\paragraph{Curse or Blessing of Dimensionality?}
Another problem with embedding is the increased dimensionality of an inverse problem.
As in NDCN, the encoder maps a scalar value to a vector of dimension $d$.
So instead of fitting a shared scalar field as in Eq~\eqref{eq:shared_operator},
we need to learn the $d$-dimension vector field.
At the same time, we have only one time series of dimension $n$ as the training data.
When $d$ is too large relative to $n$, overfitting is likely to occur.
This is illustrated in Fig~\ref{fig:line_picture}. 

One may argue that embedding is routinely used by neural ODEs in classification/regression tasks~\cite{chen2018neural,chamberlain2021grand},
and there appear to be no issues.
In classification/regression tasks, 
neural ODEs produce a flow $\Phi$ and a return time $T$ 
such that $\Phi(T, \cdot)$ 
serves as an approximator of the homeomorphism between inputs and label encoding.
In such a scenario, it is well established that lifting the dimension of the input features
enlarges the set of valid evolution functions, 
making a ``difficult" homeomorphism easier to fit \cite{dupont2019augmented}.
However, it is unnecessary to consider whether the flow $\Phi$ itself is ``correct" or not.
With that said, superfluous dimensions may still lead to unstable fit~\cite{massaroli2020dissecting}.

In NDCN, the additional embedding dimension does provide additional expressive power that compensates for
the limitation of the graph Laplacian aggregation function.
But we argue that if the graph Laplacian is too restrictive for the task, 
we should use a more general aggregation function.

\section{Dy-Net Neural Dynamics}
\label{sec:dnnd}

In the previous section, 
we have seen that an embedding-based neural dynamical model
can learn a spurious flow that fits the 
training data well, but is qualitatively different from the 
true dynamic in terms of long-term properties.

If it is so difficult to learn an encoder/decoder pair that guarantees topological conjugacy, 
do we need latent embedding in the first place? 
In this section, we show the possibility of constructing neural networks that directly model the flow in 
its original state space.

\subsection{Replacing Graph Laplacian with Permutation Invariance Aggregation}
If we build a neural network to work directly on the original state space,
one challenge is the limited expressive power of the graph Laplacian.
The graph Laplacian and its variants play a central role in graph convolution and 
graph neural networks \cite{kipf2016semi,velivckovic2017graph}.
However, the graph Laplacian is a linear aggregation operator 
that cannot handle potentially nonlinear coupling in network dynamics.
In other words, to avoid encoders and decoders, we also have to abandon the graph Laplacian and seek a more general aggregation scheme.
We propose a model based on the separation of self-dynamic and coupling operators.

\subsection{Dy-Net Neural Dynamics}
Our first approximation is to replace $\Psi(x_i(t), \{ x_j(t) : A_{ij}=1 \})$ with the sum of a node operator and an edge operator:
\begin{equation}
    \dot{x}_i(t) =  \Psi_N(x_i(t)) + \Psi_E(\{ x_j(t) : A_{ij}=1 \}).
\end{equation}

Then by Theorem 2 in \cite{zaheer2017deep}, we know that $\Psi_E(\{ x_j(t) : A_{ij}=1 \}$ can be represented as
$\rho\left(\sum_{j :A_{ij}=1 } \phi(x_{j})\right)$, where $\rho$ and $\phi$ are two scalar transformations.

Inspired by the pairwise coupling feature in \cite{chen2014unsupervised}, 
we further assume that $\rho$ is additive and that each $x_j$ can have a non-linear coupling effect on $x_i$ governed by a shared 
term $\Psi_e(x_i(t),x_j(t))$.
This leads to the following sum decomposition of the operator $\Psi$:
\begin{equation}
    \label{eq:sum_decomposition}
    \dot{x}_i(t) =  \Psi_N(x_i(t)) + \sum_{A_{ij}=1}\Psi_e(x_i(t),x_j(t)) ). 
\end{equation}
This turns out to be the same vector-field structure that appears in \cite{barzel2013universality,barzel2015constructing,gao2016universal}.

Note that Eq~\eqref{eq:sum_decomposition} is more general than the graph Laplacian aggregation in Eq~\eqref{eq:laplacian_1d}.
It is also more general than the three-term structure in \cite{barzel2015constructing}:
\begin{equation}
    \dot{x}_i(t) =  M_0(x_i(t)) + \sum_{A_{ij}=1}M_{1}(x_i(t))M_{2}(x_j(t)) ).
\end{equation}

Eq~\eqref{eq:sum_decomposition} suggests that we can use two separate neural networks $F_\theta$ and $G_\theta$ 
to parametrise $\Psi_N$ and $\Psi_e$ (see Fig.~\ref{fig:architecture}).
This is in contrast to the single-neural-network solution in NDCN.
Compared with an encoder-decoder-based model, a model specified by Eq~\eqref{eq:sum_decomposition} offers important benefits:
The neural networks for $\Psi_N$ and $\Psi_e$ are independent of $\bm{A}$.
That means that these two neural networks $F_\theta, G_\theta$ can 
be trained even when the adjacency matrix $\bm{A}$ changes within a time series.
Dynamical network rewiring is common in many applications, such as biological and social systems.
Because of this decoupling of $\Psi_e$ from $\bm{A}$, 
we call our two-neural-network model \emph{Dy-Net Neural Dynamics} (DNND).

Fig~\ref{fig:architecture}(b) shows the architecture of DNND, which renders the following flow on networks:
\begin{equation}
   \dot{{x}}_i(t)
=F_\theta\left({x}_{i}(t)\right)+\sum_{j = 1}^{N} A_{i j} G_\theta\left({x}_{i}(t), {x}_{j}(t)\right),   
\end{equation}
for all $1\leq i \leq n$.
Clearly $x_i(t)$ is always a flow by definition.

\paragraph{Affine Regularisation.}
It is well known that neural ODE training can be numerically challenging due to approximation errors in numerical integration procedures and stiffness in the system \cite{kim2021stiff}.
To alleviate the challenge of unstable model training,
we model $F_\theta$ and $G_\theta$ as 
\begin{align}
F_\theta (x) &=  c_{f0} + c_{f1} \cdot x + f_{\theta}(x)     \\
G_\theta (x_1, x_2) &= c_{g0} + c_{g11} \cdot x_1 + c_{g12} \cdot x_2 + g_{\theta}(x_1, x_2)    
\end{align}
where $c_{f0},c_{f1}, c_{g0}, c_{g11}, c_{g12}$ are scalar parameters to capture the dominating linear terms, 
and $f_{\theta}$ and $g_{\theta}$ are MLPs to capture higher-order nonlinear terms.
This is similar to the set-up of LassoNet \cite{lemhadri2021lassonet}.
It introduces the inductive bias of preferring affine mappings.
It allows us to regularise $f_{\theta}$ and $g_{\theta}$ to promote sparse and interpretable models.
Note that the graph Laplacian is now a special case with only two nonzero terms $c_{g11}=1$ and $c_{g12}=-1$.

\subsection{Robust Fitting of Two Neural Networks}

\begin{figure}[t]
     \centering
     \begin{subfigure}[b]{0.49\linewidth}
         \centering
            \includegraphics[width=1\linewidth]{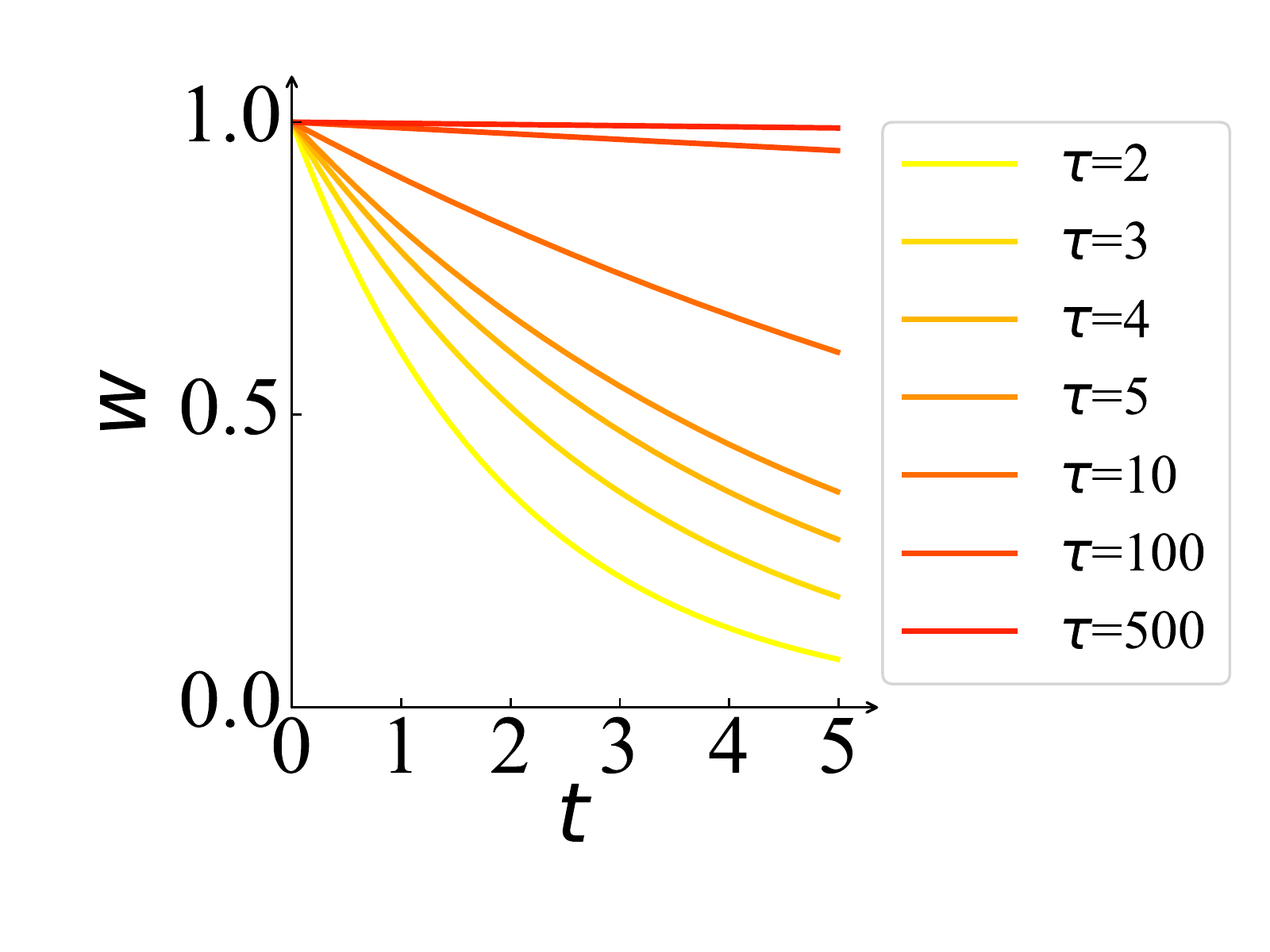}
         \caption{weight $w_k(t)$ at different temperature $\tau_k$}
         \label{fig:five over x}
     \end{subfigure}
     \begin{subfigure}[b]{0.49\linewidth}
         \centering
            \includegraphics[width=1\linewidth]{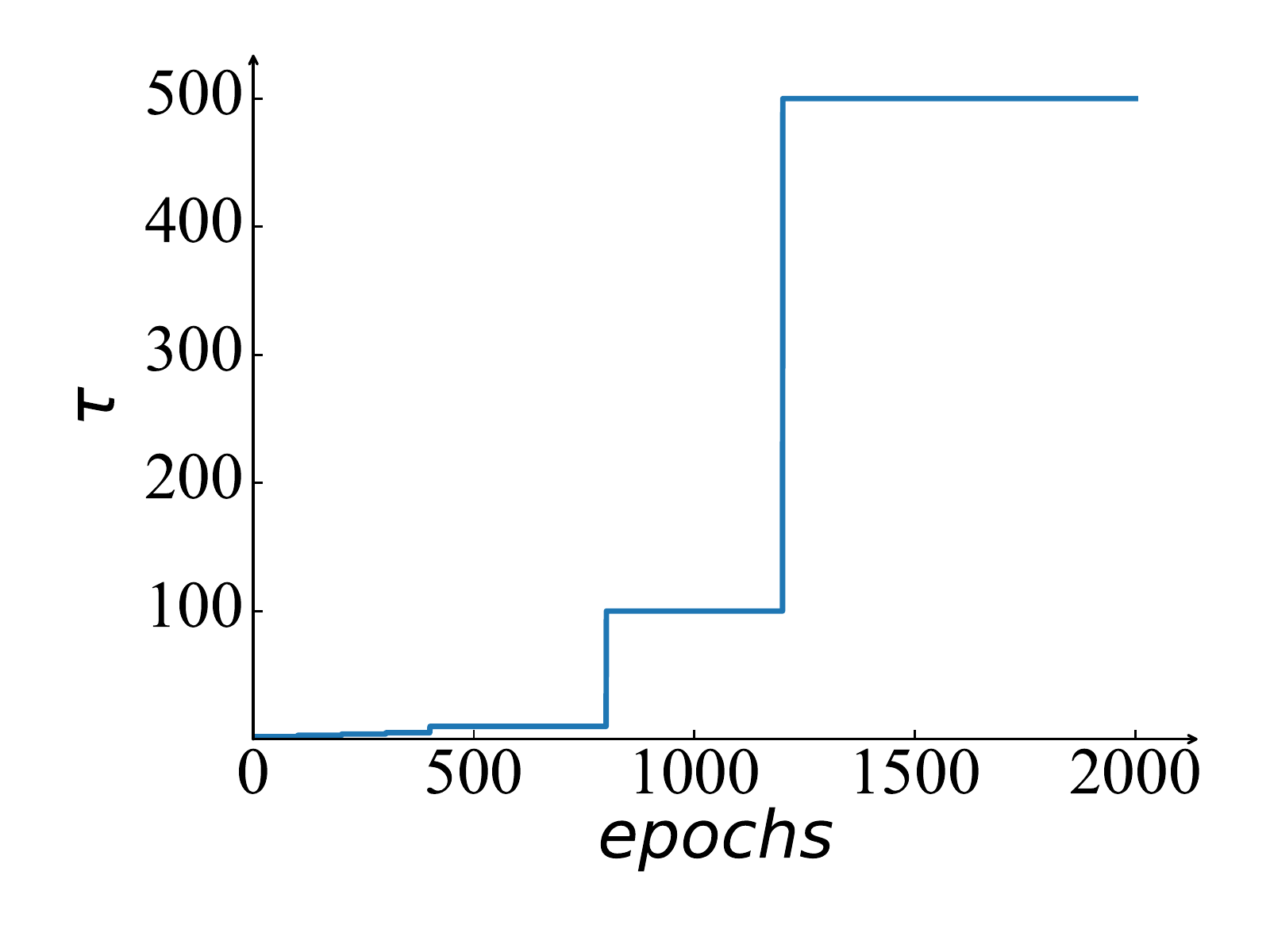}
         \caption{temperature $\tau_k$ at different epoch $k$}
         \label{fig:T of weight}
     \end{subfigure}
     
        \caption{Loss warm-up schedule for learning $F$ and $G$.
        (a) At the early stage of the training process, the contribution of the data points far away from the initial time ($t$ being large) are dampened to stabilise the training. 
        (b) A temperature variable $\tau$ controls how the dampening is relaxed with the number of epochs.
        The temperature increases every 400 epochs.
        }
        \label{weighting schedule}
\end{figure}

Given a time series $D = \bm{x}(t_1), \bm{x}(t_2), \dots, \bm{x}(t_N)$,
the learning problem is to find 
$F_\theta: \mathbb{R} \rightarrow \mathbb{R}$ and 
$G_\theta: \mathbb{R}^2 \rightarrow \mathbb{R}$
that minimise the following loss:
\begin{equation}
     \mathcal{L}= \mathcal{R}(\hat{\bm{x}}(t), D). 
\end{equation}

To reduce the impact of numerical error accumulation from long-range integration \cite{Milani1987IntegrationEO},
we devise a warm-up schedule to adapt the loss function dynamically.
At the early stage of training, 
the model parameters are strongly affected by time-series data points closer to the initial value, which results in smaller integration errors.
More specifically, we adopt a weighting schedule for the loss function with evolving weight function $w_k$ at epoch $k$ as follows:
\begin{equation}
    w_k(t) = e^{-t/\tau_k }, 
\end{equation}
where $\tau_k$ monotonically increases with the number of training epochs $k$.
So the loss function evolves as follows:
\begin{equation}
     \mathcal{L}_k= \sum_{y_t\in D}\mathcal{R}(\hat{\bm{x}}(t), y_t) \cdot w_k(t)
\end{equation}

\section{Experiments} 
We use several approaches to recover evolution operators from time series and then compare the 
quality of the recovered models.
We closely follow \cite{zang2020neural} to set up the ground-truth models and to simulate the observation data.
In this setup, a ground-truth model is a coupled system of ODEs defined on a random network.
From initial conditions, we simulate a time series as the observation data.
Our goal is to recover a neural-network model for the evolution operators in the ground-truth ODE system.

\subsection{Ground-truth Dynamical Models and Observation Data}
We considered an extensive set of linear and nonlinear ODEs (Table~\ref{tab:dynamics}) on different complex networks, including a fixed grid network from~\cite{zang2020neural} 
and random networks based on the Erd\'os and R\'enyi (ER) model~\cite{gilbert1959random}, 
the Albert-Barab\'asi (BA)  model~\cite{barabasi1999emergence}, 
the Watts-Strogatz (WS) model~\cite{watts1998collective},  
and the LFR model~\cite{fortunato2010community,lancichinetti2008benchmark}.

\begin{table}[t]
    \centering
    \begin{tabular}{ll}\toprule
        dynamic & evolution rule for $x_i$ \\ \midrule
        Heat & $
        \dot{x}_i
        = - 0.1
        \sum_j A_{i j}
        (x_{i} - x_{j}) $ \\
         Biochemical & $
        \dot{x}_i
        =
        1
        - 
        0.1
        x_{i}- 
        0.01
        \sum_j A_{i j}
        x_{i} x_{j} $ \\
        BirthDeath &  $
        \dot{x}_i
        = -0.1 
        x_{i}^2
        +
        0.2
        \sum_j
        A_{i j} 
        x_{j}
        $ \\ \bottomrule
    \end{tabular}
    \caption{Dynamical systems for generating observation time-series. 
    More background on these formulas can be found in ~\protect\cite{barzel2013universality}.
    }
    \label{tab:dynamics}
\end{table}

\paragraph{Training Time Series}
We follow the setup in \cite{zang2020neural} and 
randomly sample $80$ (irregularly spaced) times $0 \leq t_1 < t_2 < \dots < t_{80} \leq 5$
for training.
The initial conditions $\bm{x}(0)$ 
of the dynamical variables on vertices are set with random values from $[0,25]$.
We use the Runge–Kutta–Fehlberg (RKF45) solver to generate $\bm{x}(t_1), \bm{x}(t_2), \dots, \bm{x}(t_{80})$ as the training time series.

\begin{figure}[b]
    \centering
       \hspace*{\fill}
    \begin{subfigure}[b]{0.45\linewidth}
         \centering
           \includegraphics[width=\linewidth]{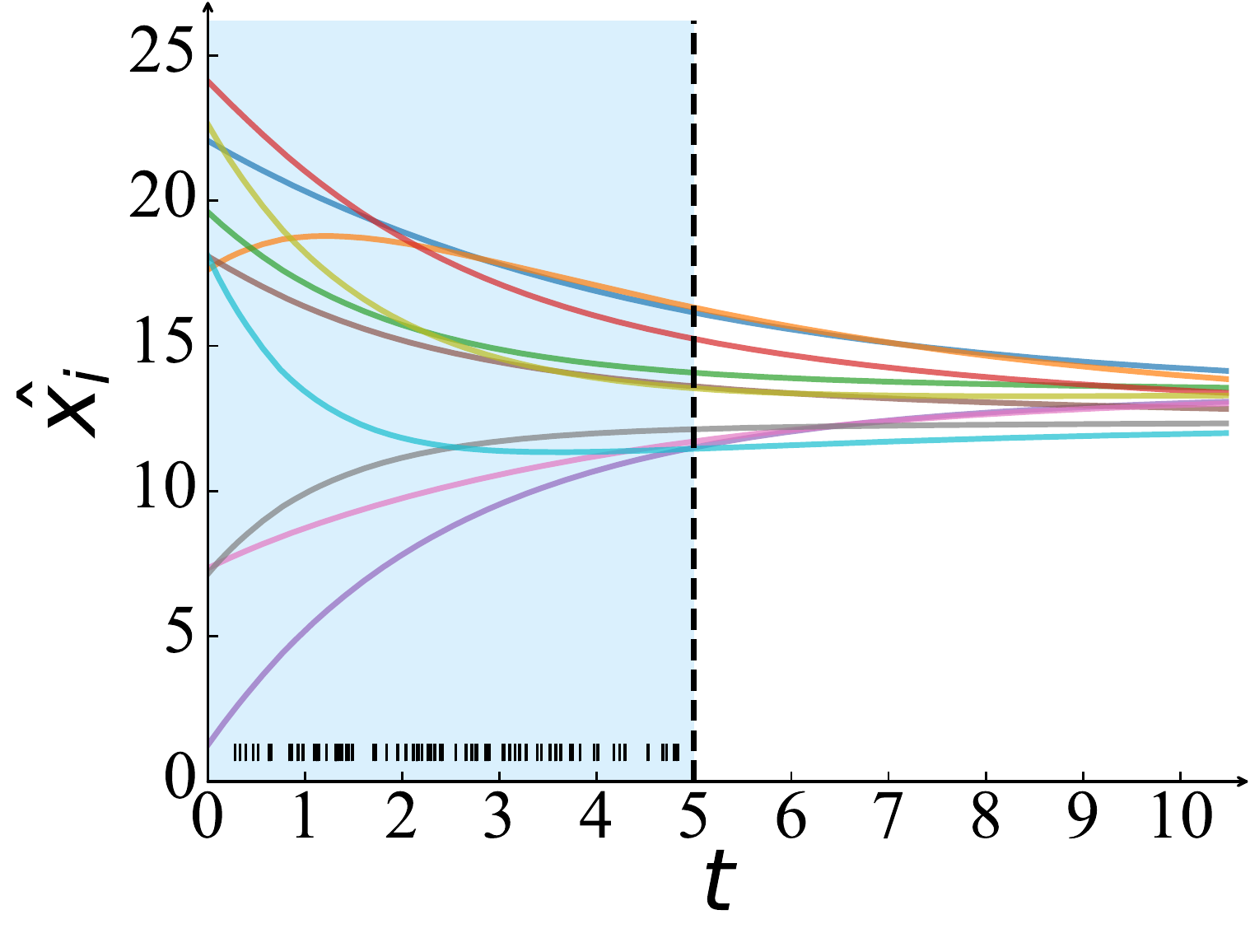}
         \caption{estimated trajectories}
         \label{fig:model-x-t}
     \end{subfigure}
     \hspace*{\fill}
     \begin{subfigure}[b]{0.45\linewidth}
         \centering
         \includegraphics[width=\linewidth]{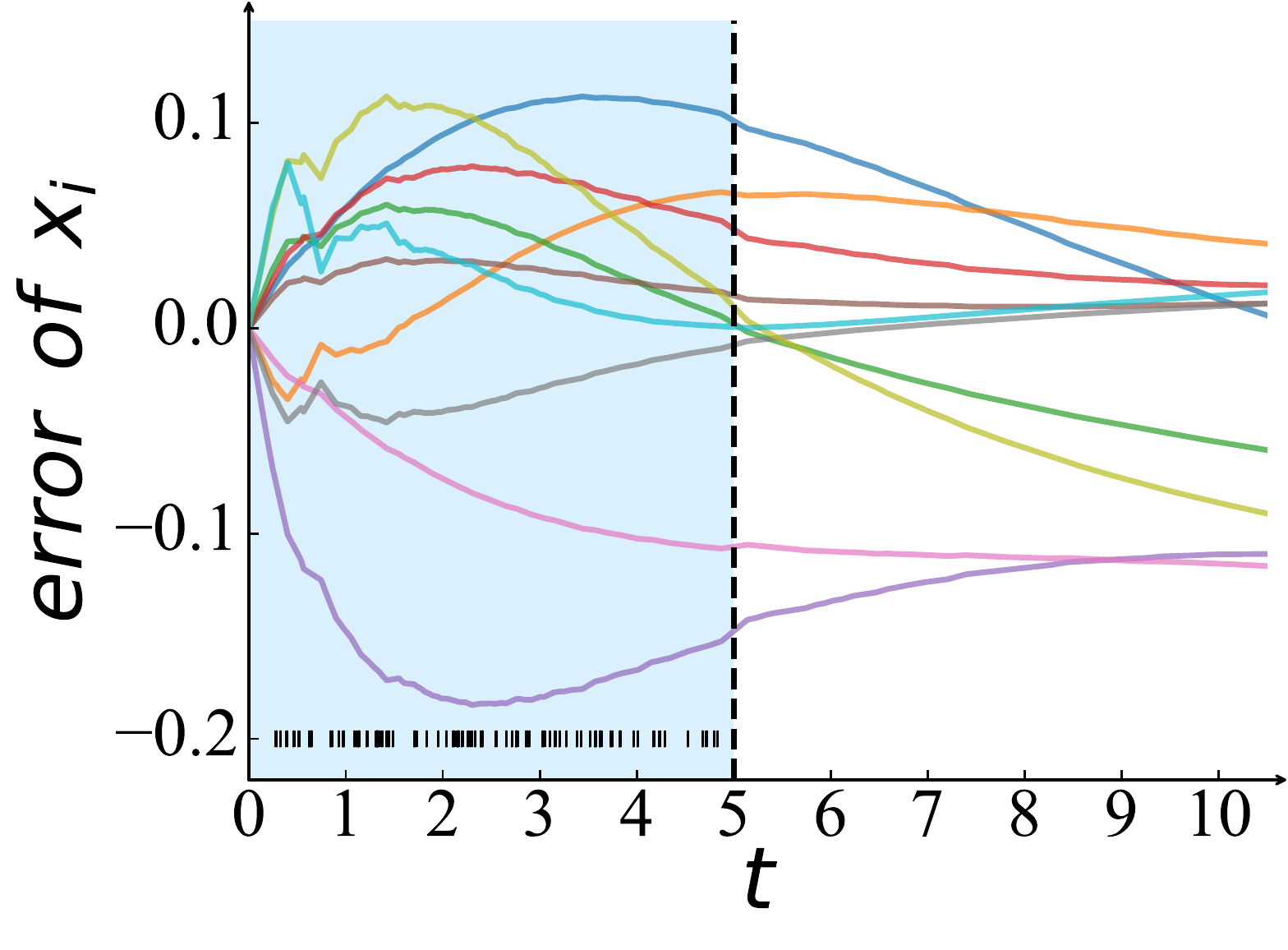}
         \caption{estimation error}
         \label{fig:diff of model and true x-t}
     \end{subfigure}
     \hspace*{\fill}
    \caption{Time series estimated by DNND and the estimation error on the same dataset in Fig~\ref{fig:ndcn_overfitting}. The estimation error is bounded. 
    (Note that the y axis here has a different range.)}
    \label{fig:fitted_trajectories_DNND}
\end{figure}

\subsection{Generalisation beyond the Training Data}

Table~\ref{tab:extrapolation until steady state} compares the prediction performance from the short-term prediction to the long-term prediction.
One can see that NDCN performs reasonably well near the training data, but its long-term prediction is unreliable,
with the mean absolute percentage error (MAPE) even exceeding $100\%$.
In contrast, DNND produces accurate predictions, even well beyond the  region of training data.
The pattern persists on other network topologies (see more results in the supplementary material).

\begin{table}[!t]
    \centering
   \resizebox{\linewidth}{!}{
    \begin{tabular}{lrrr}
        \toprule
         dynamics & time & NDCN & DNND \\ \midrule
         \multirow{3}{*}{heat} & 0-5 & 1.3$\pm$0.4 & \pmb{0.3$\pm$0.1} \\
         ~ & 5-6 & 1.5$\pm$0.4 & \pmb{0.5$\pm$0.3} \\
         ~ & 40-50 & 322.4$\pm$148.0 & \pmb{11.9$\pm$1.8} \\ \midrule
        \multirow{3}{*}{biochemical} & 0-5 & 6.2$\pm$0.8 & \pmb{1.3$\pm$0.2} \\
         ~ & 5-6 & 12.5$\pm$1.8 & \pmb{1.0$\pm$0.1}\\
         ~ & 40-50 & 27142.4$\pm$29581.0 
         & \pmb{46.0$\pm$11.9} \\ \midrule
        \multirow{3}{*}{birthdeath}  & 0-5 & 1.4$\pm$0.3 & \pmb{0.7$\pm$0.3} \\
         ~ & 5-6 & 1.5$\pm$0.2 & \pmb{0.3$\pm$0.1} \\
         ~ & 40-50 & 1673.4$\pm$900.8 & \pmb{0.4$\pm$0.1}\\\bottomrule
    \end{tabular}
    }
    \caption{Forecast MAPE (\%) on the grid network. 
    Models are trained on 80 observations from $[0,5]$ and evaluated at random times in three periods:
    interpolation [0-5], short-term [5-6], and long-term [40-50].
    }
    \label{tab:extrapolation until steady state}
\end{table}

\begin{figure}[t]
    \centering
    \begin{subfigure}[b]{0.49\linewidth}
         \centering
            \includegraphics[width=1\linewidth]{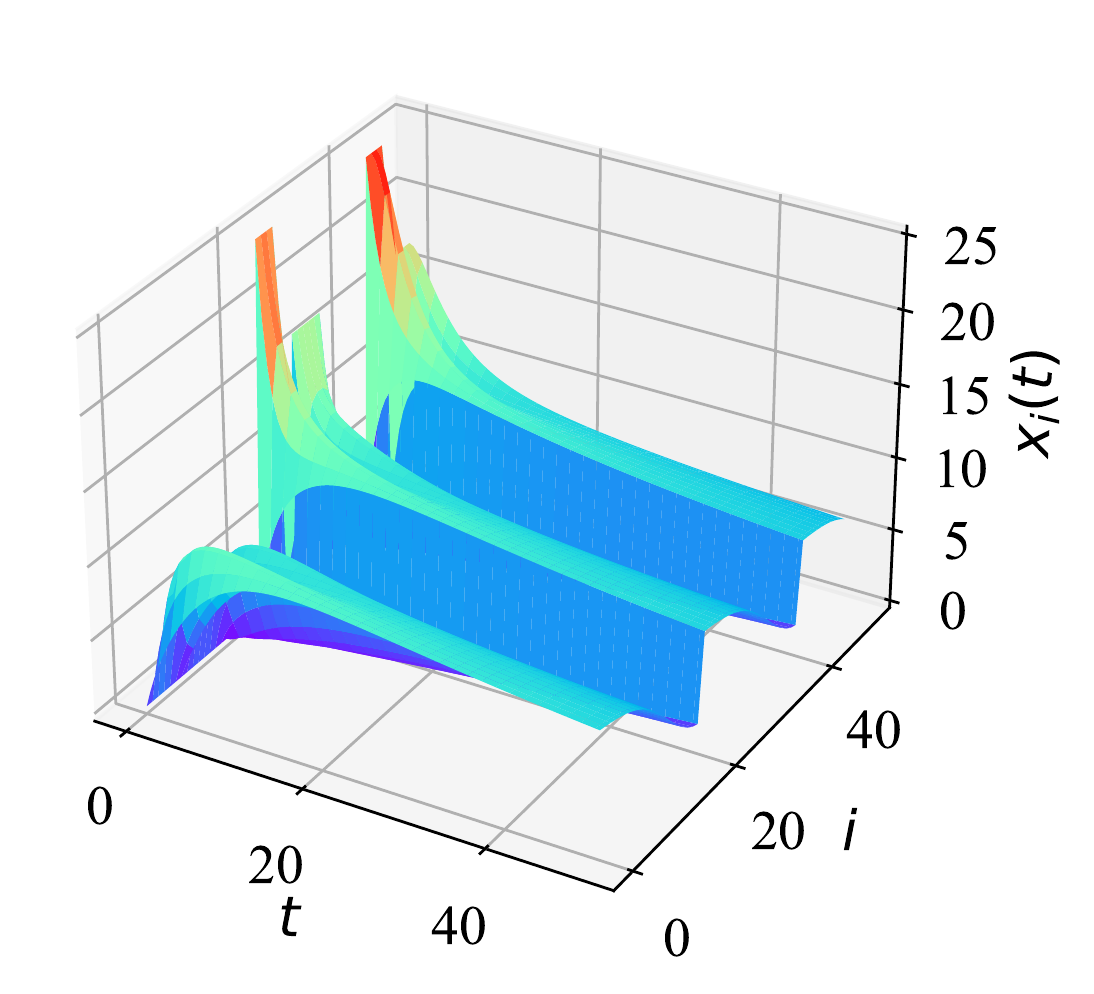}
         \caption{ground-truth}
         \label{fig:steady heat grid true 50}
     \end{subfigure}
     \begin{subfigure}[b]{0.49\linewidth}
         \centering
            \includegraphics[width=1\linewidth]{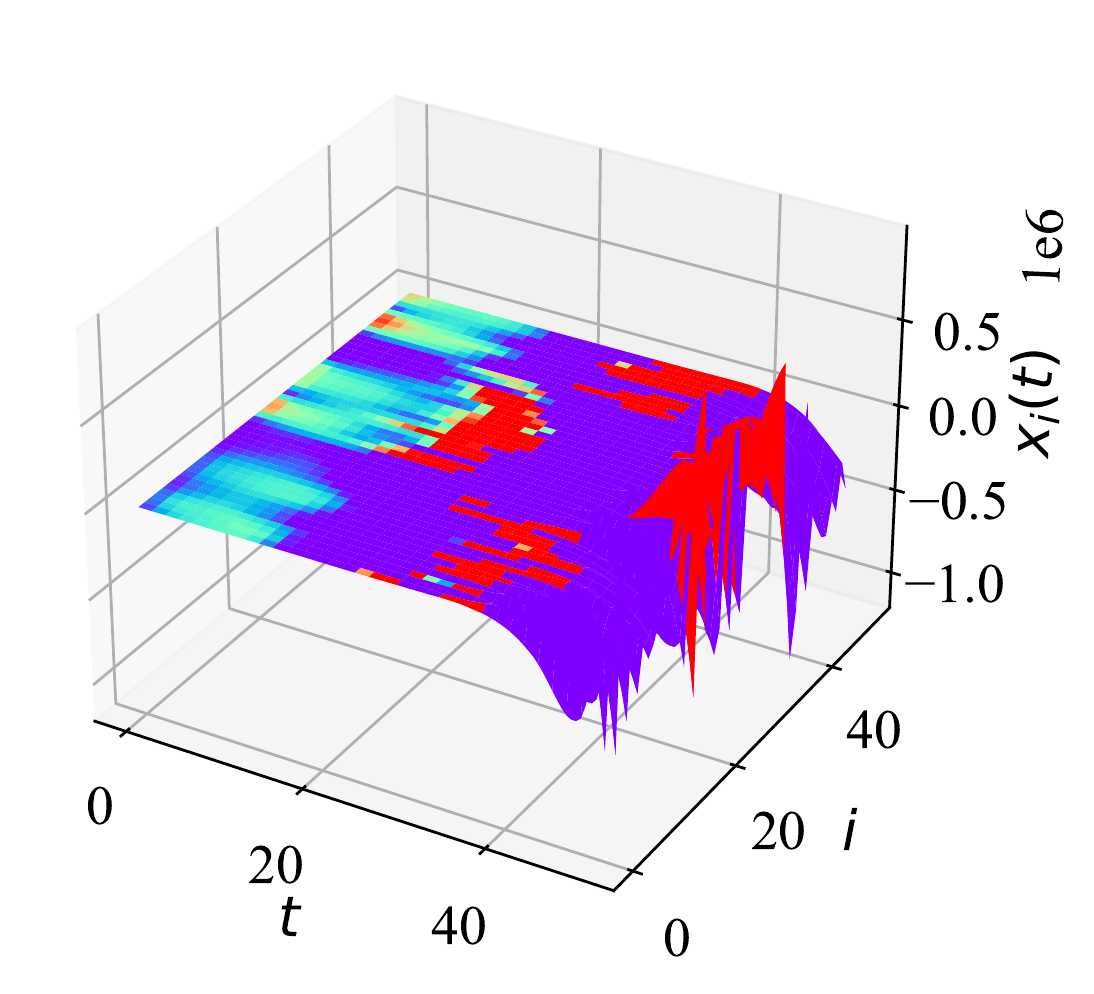}
         \caption{NDCN predictions}
         \label{fig:ndcn steady heat grid predict 50}
     \end{subfigure}
     \begin{subfigure}[b]{0.49\linewidth}
         \centering
            \includegraphics[width=1\linewidth]{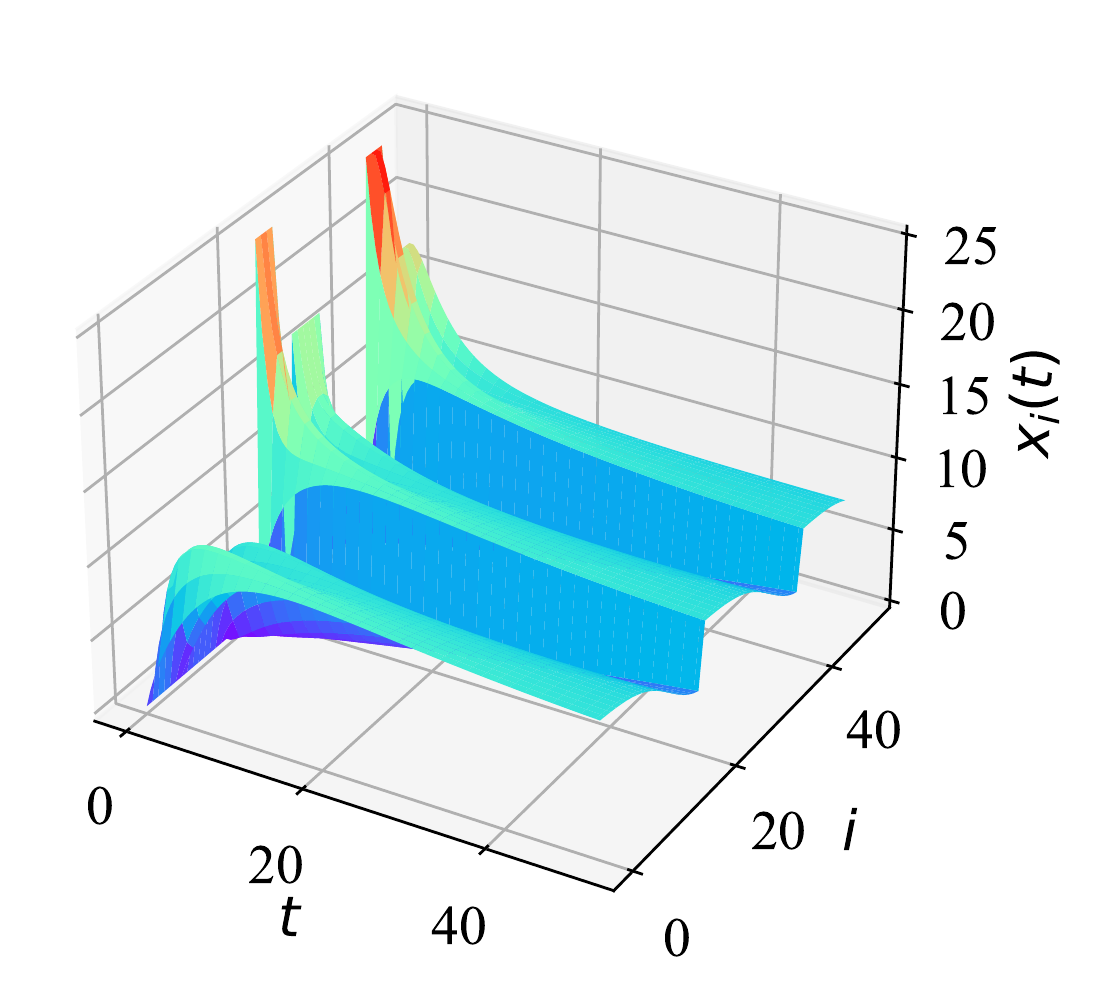}
         \caption{DNND predictions}
         \label{fig:steady heat grid predict 50}
     \end{subfigure}
       \begin{subfigure}[b]{0.49\linewidth}
         \centering
            \includegraphics[width=1\linewidth]{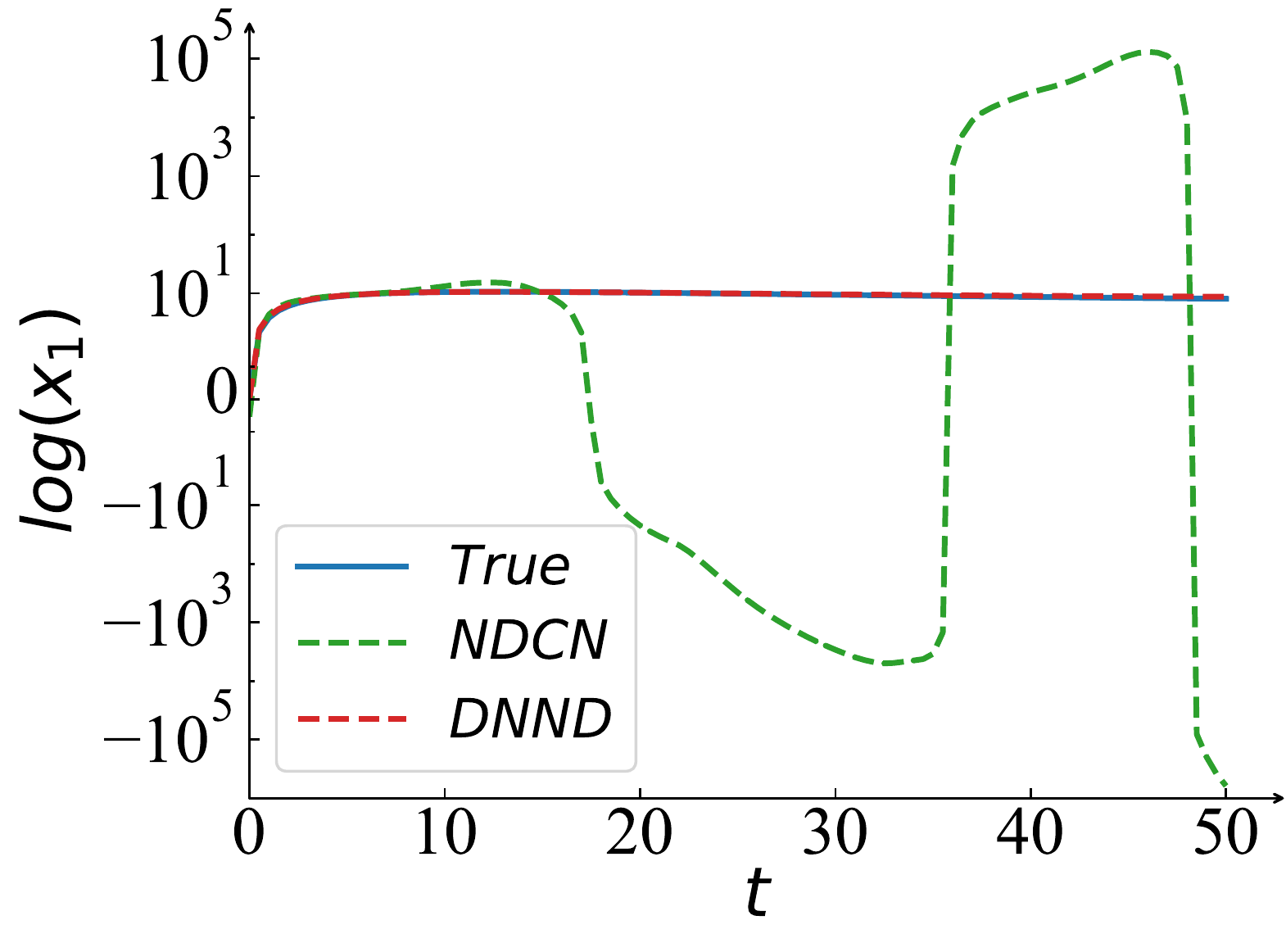}
         \caption{one vertex}
         \label{fig: mix dnnd and ndcn steady heat grid one node}
     \end{subfigure}
    \caption{Predicted time series on 50 vertices. 
    NDCN predictions are completely wrong in the long term while DNND predictions are consistent with the ground truth.
    Note in (b) the vertical axis has a wider range.
    (d) Time series for node $x_1$; symmetric log plot~\protect\cite{danz2021ultrafast} was used due to the large difference in scales.} 
    \label{fig:steady state on heat gridx_}
\end{figure}

Fig.~\ref{fig:steady state on heat gridx_} shows time series on 50 vertices.
NDCN produces time series with values far exceeding the physically reasonable values.

\subsection{Recovery of Dynamic}

The overfitting of NDCN is also revealed in the largest Lyapunov exponent of the fitting model.
Table~\ref{tab:lyapunovExponent} shows that all dynamics are stable with the grid network, with the 
negative Lyapunov exponents. 
The DNND estimates are similar to the true values. 
In contrast, NDCN produces positive Lyapunov exponents, suggesting unstable dynamics.

\begin{figure}[hpt]
    \centering
    \begin{subfigure}[b]{0.32\linewidth}
         \centering
         \includegraphics[width=\linewidth]{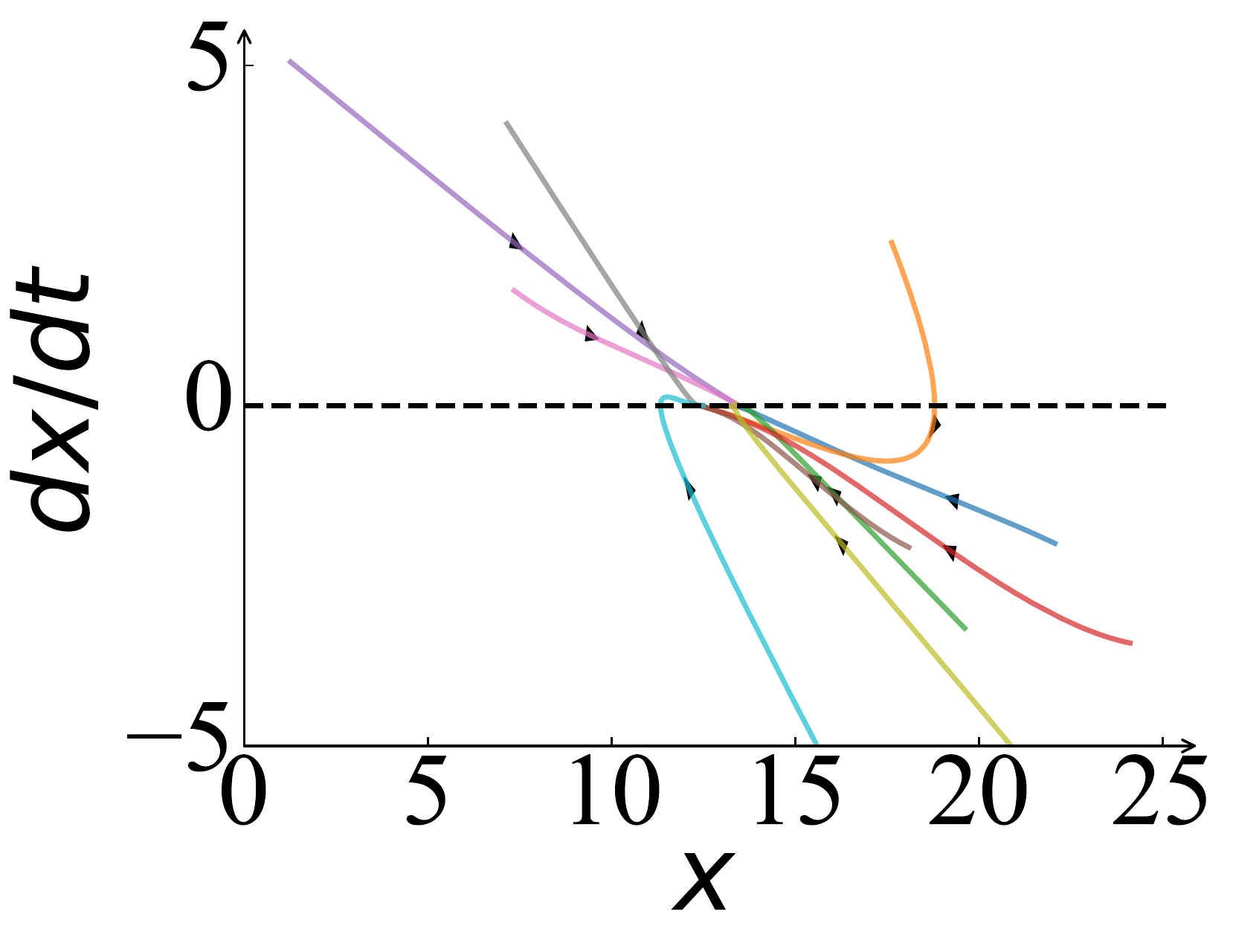}
         \caption{ground truth}
     \end{subfigure}
     \hspace*{\fill}
     \begin{subfigure}[b]{0.32\linewidth}
         \centering
         \includegraphics[width=\linewidth]{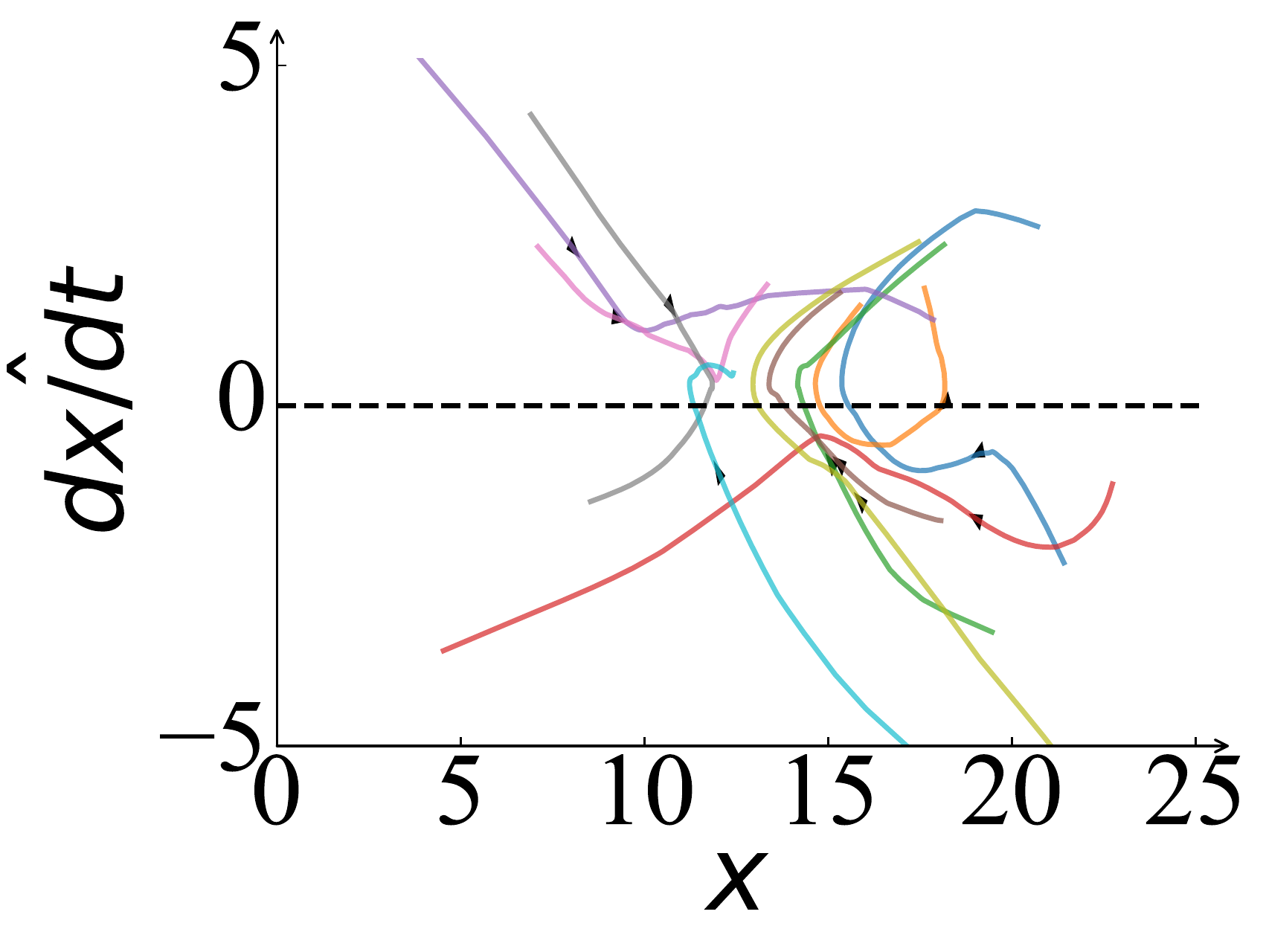}
         \caption{NDCN}
     \end{subfigure}
     \hspace*{\fill}
    \begin{subfigure}[b]{0.32\linewidth}
         \centering
         \includegraphics[width=\linewidth]{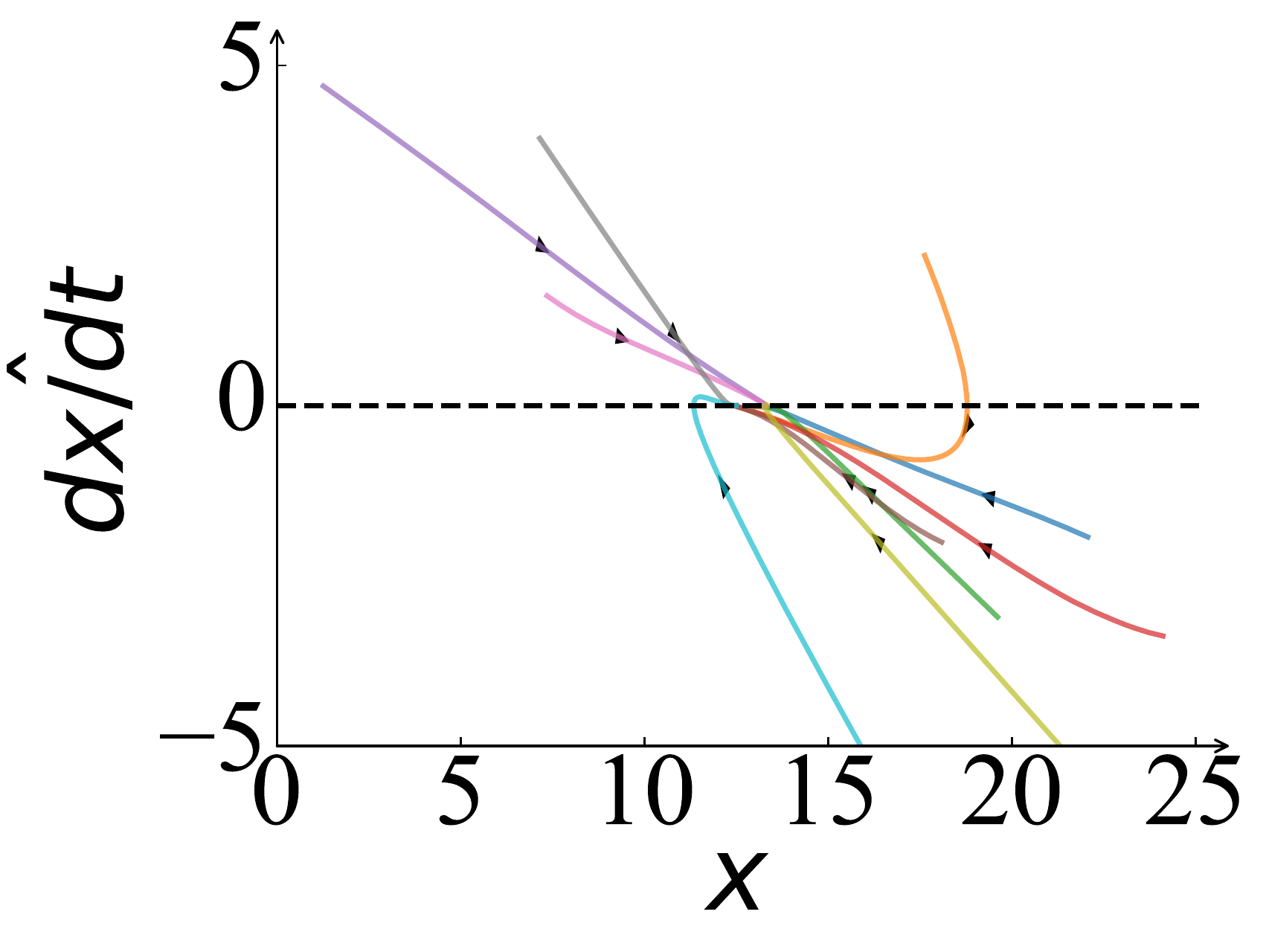}
         \caption{DNND}
     \end{subfigure}
    
    \caption{2D projection of trajectories from 9 random nodes in the grid network.
    The dynamic is the heat flow.
    The ground-truth trajectories converge to the only fixed point of the system, with every node 
    having the average heat.
    (b) NDCN trajectories are not converging to a fixed point.
     }
    \label{fig:phase_plot}
\end{figure}

\begin{table}[htb]
    \centering    
    \begin{tabular}{lrrr}
        \toprule
        dynamics & true value & NDCN & DNND \\ \midrule
        Heat & $-$35.83 & 9.36 & $-$34.97 \\ 
        Biochemical & $-$9.34 & 18.94 & $-$10.23 \\ 
        Birthdeath & $-$92.07 & 15.70 & $-$87.52   \\       \bottomrule
    \end{tabular}
    \caption{The largest Lyapunov exponent of the ground-truth models and the estimated models.
    Positive values indicate unstable (potentially chaotic) dynamics.
    }
    \label{tab:lyapunovExponent}
\end{table}

we can also see the fixed point, if any, of the fitted model using a projection into the 2D space spanned by 
$x_i$ and $\dot{x}_i$.
Fig.~\ref{fig:phase_plot} shows that DNND converges to the correct fixed point.
In contrast, NDCN does not converge at all.

\begin{figure}[htbp]
    \begin{subfigure}[b]{0.4\linewidth}
         \centering
            \includegraphics[width=1\linewidth]{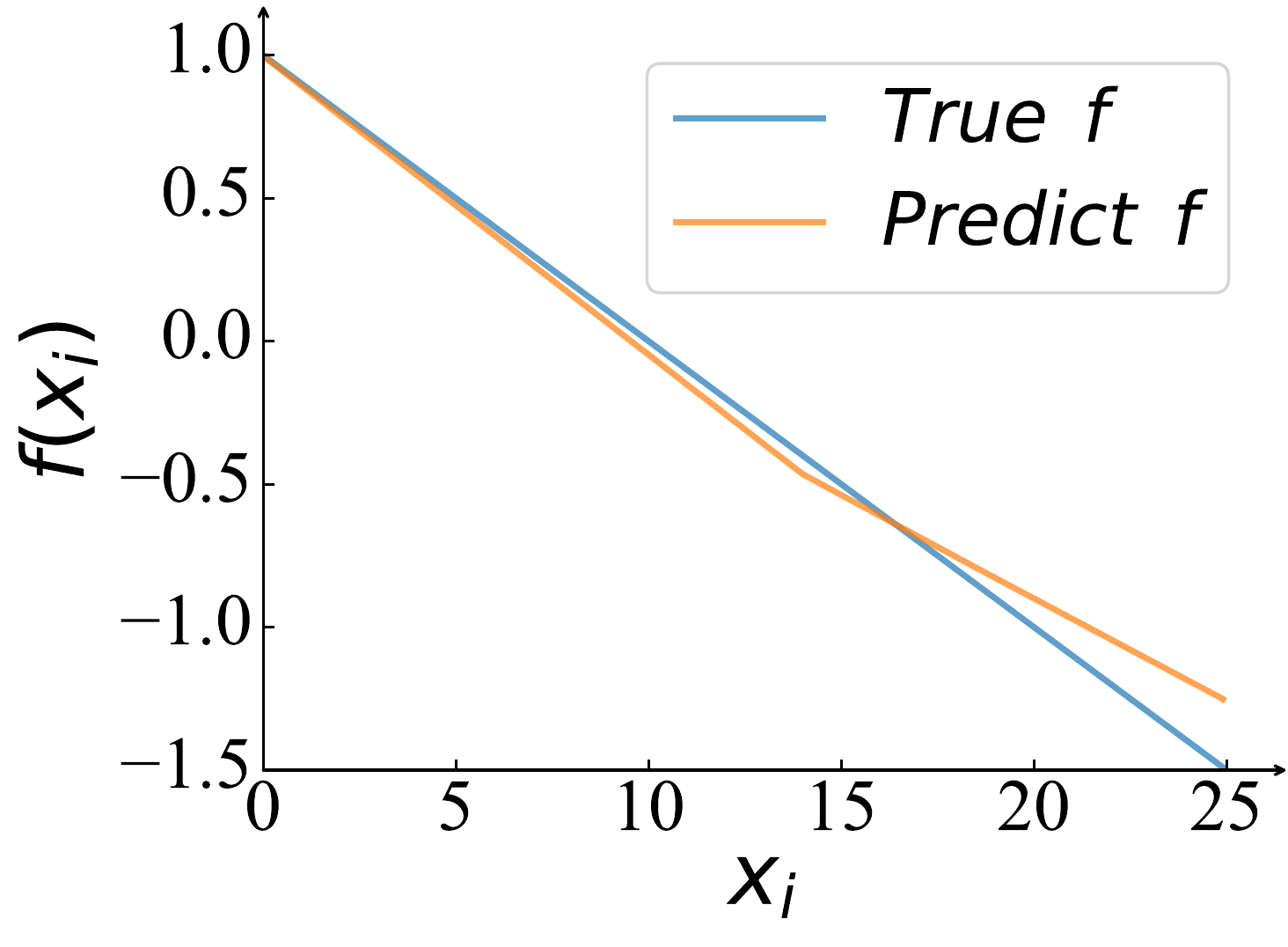}
            \caption{}
         \label{fig:Biochemical on community of F}
    \end{subfigure}
    \hspace*{\fill}
    \begin{subfigure}[b]{0.6\linewidth}
         \centering
            \includegraphics[width=1\linewidth]{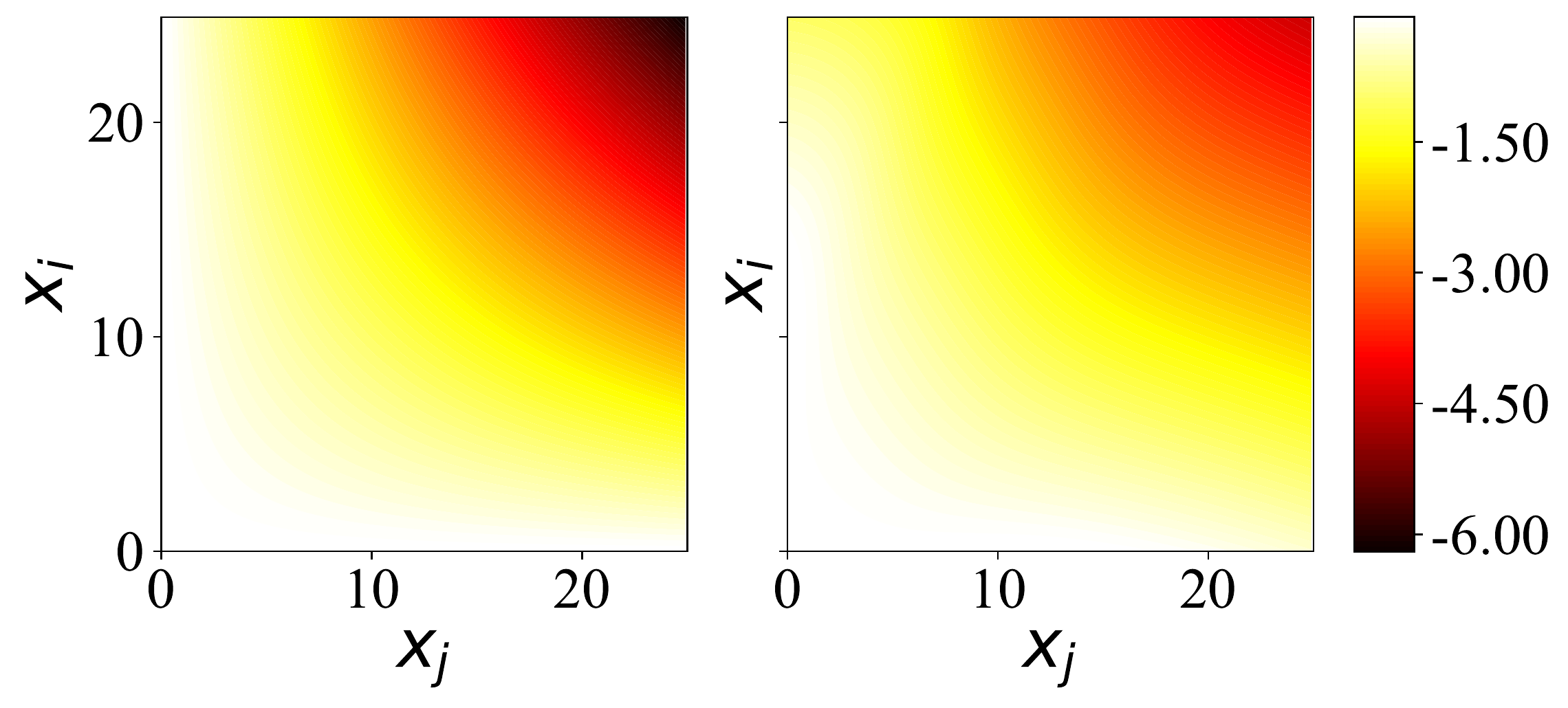}
            \caption{}
         \label{fig:Biochemical on community of true and DNND}
    \end{subfigure}
    \hspace*{\fill}
\caption{Functions recovered by DNND.
(a) true and estimated graphs of $F_{\theta}$  
(b) heatmaps for the value of $G_{\theta}(x_i, x_j)$, with the true one on the left and the estimated one on the right.
The true dynamic is the biochemical dynamic in Table~\ref{tab:dynamics}.}
\label{fig:governing_function_biochemicalmap}
\end{figure}

\paragraph{Recovering Governing Evolution Function.}
As shown in \cite{murphy2022duality}, predictability and reconstructability are independent and sometimes incompatible problem properties.
A time series with less variability is easier to predict but provides less 
information for recovering the evolution function.
Fig.~\ref{fig:governing_function_biochemicalmap} shows that from the two subnetworks in DNND, we can reasonably recover the 
vector field in terms of self-dynamic and coupling effect: $\Psi_N$ and $\Psi_e$ in Eq.~\eqref{eq:sum_decomposition}.

Note that NDCN's vector field is defined in a latent space and it is impossible to recover the evolution function in terms of the original state variables.
Unlike NDCN, which models the governing function on latent variables,
DNND directly captures the governing function of the dynamic
in terms of the original physical measurements.
From the $F_{\theta}$-network and $G_{\theta}$-network, we can recover the input-output dependency. 
It is another advantage of our embedding-free approach.

Furthermore, if $\Psi_N$ and $\Psi_e$ have analytical forms in terms of basis functions, 
then symbolic regression can potentially be applied to recover the exact algebraic forms of the two functions. 
We show some results in the Appendix.

\section{Conclusion}

We show that blindly applying  an encoder-decoder structure,
without a sound principle such as topological conjugacy,
is harmful to correct recovery of a dynamic flow from observations.
Through our proposed \emph{Dy-Net Neural Dynamics} (DNND) model, we show that
state embedding is unnecessary for a broad class of flow-recovery problems.
Empirically, DNND reliably recovers a wide range of dynamical systems on different network topologies.

As DNND models the vector field in the original state space, the fitted model can serve as a basis for 
discovering analytical, hence interpretable, differential equations using techniques such as symbolic regression~\cite{brunton2016discovering}.
We presented some early results in the Appendix.

Theoretically, it would be interesting to characterise the learnability of dynamical flows, up to topological equivalence,
from irregularly spaced observations.
This is related to finding a suspension flow under a roof function~\cite{fisher2019hyperbolic}. 
But as we have access to only samples of the homeomorphism, it is a machine learning problem.
One particularly interesting question worth exploring is
how to parametrise time changes ($\pdv{t}$) given a finite sample from an unknown roof function.

\section*{Acknowledgments}

This work was supported by the National Key R\&D Program of China under Grant Nos. 2021ZD0112501 and 2021ZD0112502; the National Natural Science Foundation of China under Grant Nos. U22A2098, 62172185, 62206105 and 62202200; Jilin Province Capital Construction Fund Industry Technology Research and Development Project No. 2022C047-1; Changchun Key Scientific and Technological Research and Development Project under Grant No. 21ZGN30;
Key Laboratory of Symbolic Computation and Knowledge Engineering of Ministry of Education Open Research Project No. 93K172021K05.

\bibliographystyle{unsrt}
\bibliography{ijcai23}

\end{document}